\documentclass[11pt]{article}

\usepackage[preprint]{acl}

\usepackage{times}
\usepackage{latexsym}

\usepackage[T1]{fontenc}

\usepackage[utf8]{inputenc}

\usepackage{microtype}

\usepackage{inconsolata}

\usepackage{graphicx}

\usepackage{amsmath}

\usepackage[linesnumbered,ruled,vlined]{algorithm2e}

\usepackage{amssymb}

\usepackage{booktabs}

\usepackage{xcolor}
\usepackage{multirow}

\usepackage[most]{tcolorbox}
\newtcolorbox{promptbox}[1][]{
  colback=gray!10!white,
  colframe=gray!60!black,
  title={#1},
  fonttitle=\bfseries,
  boxrule=0.5mm,
  sharp corners,
  breakable,
  fontupper=\ttfamily\small, 
  #1
}

\usepackage{booktabs}  
\usepackage{multirow}  
\usepackage{tabularx}  
\usepackage[table,xcdraw]{xcolor} 
\usepackage{colortbl}

\usepackage{enumitem}

\title{Beyond Monolithic Architectures: A Multi-Agent Search and Knowledge Optimization Framework for Agentic Search}

\author{
\textbf{Yiqun Chen}$^{1}$\thanks{Equal contribution.}, 
\textbf{Lingyong Yan}$^{2}$\footnotemark[1], 
\textbf{Zixuan Yang}$^{1}$\footnotemark[1], 
\textbf{Erhan Zhang}$^{1}$\footnotemark[1], 
\textbf{Jiashu Zhao}$^{2}$, \\
\textbf{Shuaiqiang Wang}$^{2}$, 
\textbf{Dawei Yin}$^{2}$, 
\textbf{Jiaxin Mao}$^{1}$\thanks{Corresponding author.} \\
$^{1}$ Renmin University of China \\
$^{2}$ Baidu Inc. \\
\texttt{chenyiqun990321@ruc.edu.cn}, \texttt{maojiaxin@gmail.com}
}

\begin{document}
\maketitle
\begin{abstract}
Agentic search has emerged as a promising paradigm for complex information seeking by enabling Large Language Models (LLMs) to interleave reasoning with tool use. 
However, prevailing systems rely on monolithic agents that suffer from structural bottlenecks, including unconstrained reasoning outputs that inflate trajectories, sparse outcome-level rewards that complicate credit assignment, and stochastic search noise that destabilizes learning. 
To address these challenges, we propose \textbf{M-ASK} (Multi-Agent Search and Knowledge), a framework that explicitly decouples agentic search into two complementary roles: Search Behavior Agents, which plan and execute search actions, and Knowledge Management Agents, which aggregate, filter, and maintain a compact internal context. 
This decomposition allows each agent to focus on a well-defined subtask and reduces interference between search and context construction. 
Furthermore, to enable stable coordination, M-ASK employs turn-level rewards to provide granular supervision for both search decisions and knowledge updates. 
Experiments on multi-hop QA benchmarks demonstrate that M-ASK outperforms strong baselines, achieving not only superior answer accuracy but also significantly more stable training dynamics.\footnote{The source code for M-ASK is available at https://github.com/chenyiqun/M-ASK.}
\end{abstract}

\section{Introduction}
\label{sec:intro}

The rapid evolution of Large Language Models (LLMs) has fundamentally reshaped information retrieval, driving a paradigm shift from passive keyword matching to \textit{Agentic Search}~\cite{shi2025deepresearchsystematicsurvey}. Unlike traditional retrieval systems, agentic search systems function as autonomous decision-makers capable of iterative planning, external tool querying, and information synthesis to address complex, multi-hop user needs. Represented by Search-r1~\cite{jin2025search}, recent advancements~\cite{song2025r1,song2025r1++,zheng2025deepresearcher} integrate reasoning directly into LLM-based multi-round search workflows. By executing multi-step search within a single response via end-to-end optimization, these approaches significantly enhance performance on complex QA tasks—capabilities that remain beyond the reach of static search paradigms~\cite{ma2023query,ke2024bridging}.

\begin{figure}[t]
  \centering
  \includegraphics[width=0.49\textwidth]{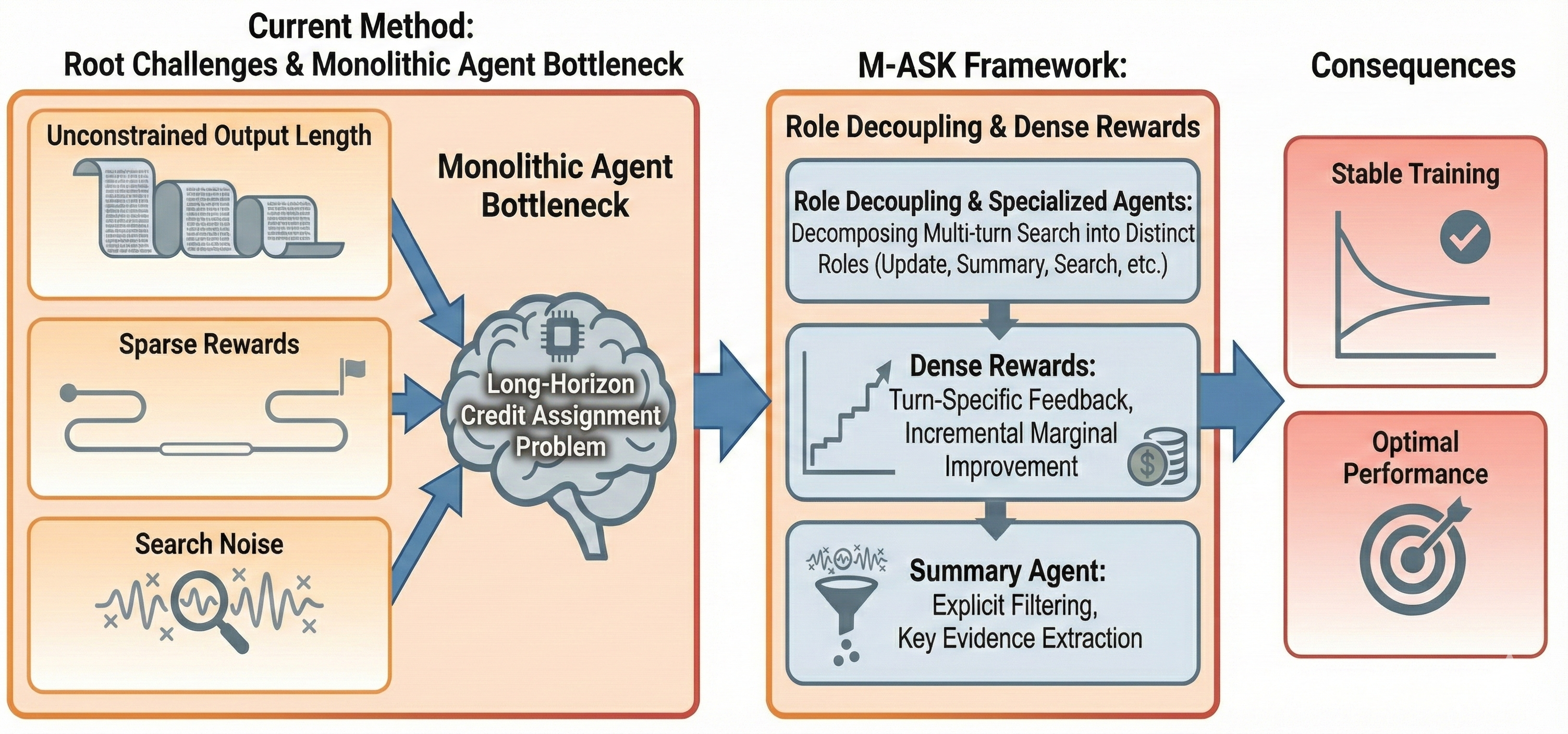}
  \caption{Challenges of current monolithic methods and the M-ASK solution. Existing agents struggle with the \textit{long-horizon credit assignment problem} caused by unconstrained output length, sparse rewards and search noise. M-ASK addresses these bottlenecks through role decoupling and turn-level dense rewards.}
  \label{fig:challenges}
\end{figure}

However, training robust agents to navigate these dynamic search environments presents significant challenges. Prevailing approaches predominantly adopt a \textit{monolithic} architecture that executes multi-round search within a single, continuous response. In this paradigm, the LLM shoulders the heavy burden of both trajectory planning and information processing at every iterative step of the generation. We argue that this monolithic design is structurally vulnerable to three intertwined obstacles: (i) \textbf{unconstrained output length}, where agents generate verbose reasoning chains that extend search horizons without necessarily increasing information density; (ii) \textbf{sparse rewards}, where feedback is typically delayed until task completion, hindering effective step-wise credit assignment; and (iii) \textbf{search noise}, where external tools such as search engines introduce noise and irrelevant data into the context.

As illustrated in Figure~\ref{fig:challenges}, these challenges are not isolated; rather, they compound to destabilize training. The interaction between extended trajectories and sparse, outcome-only feedback creates a severe \textit{long-horizon credit assignment} problem: optimization algorithms struggle to attribute the final reward to specific, distant tokens. This fragility is further exacerbated by search engine noise—when stochastic context infiltrate an already lengthy and sparsely rewarded episode, the learning signal becomes effectively indistinguishable from variance. Consequently, monolithic agents~\cite{jin2025search} frequently suffer from suboptimal state and high training instability~\cite{deng2025grpo}.

To overcome the limitations of the monolithic method, we propose \textbf{M-ASK} (\textbf{M}ulti-\textbf{A}gent \textbf{S}earch and \textbf{K}nowledge), a framework that fundamentally disentangles the decision-making of search from the burden of information integration. 
Rather than relying on a single agent, M-ASK orchestrates a collaboration between two specialized roles:
\begin{enumerate}[leftmargin=*, itemsep=2pt, topsep=2pt, parsep=0pt]
    \item \textbf{Search Behavior Agents} (including the Planning, Search, and Answer Agents), which focus exclusively on trajectory planning, interacting with search engine, and generating answers;
    \item \textbf{Knowledge Management Agents} (including the Summary and Update Agents), which act as dynamic filters to prune noisy observations, update and maintain a concise internal knowledge state, and thus, constrain context length.
\end{enumerate}
Crucially, M-ASK abandons the reliance on sparse feedback in favor of \textbf{turn-specific dense rewards}. By jointly optimizing all agents with immediate, turn-aware supervision, we ensure that the Knowledge Management Agents actively stabilize the state space, empowering the Search Agents to conduct more accurate planning. Consequently, this synergistic collaboration effectively mitigates the impact of noise and long horizons.

Our contributions are summarized as follows:
\begin{itemize}[leftmargin=2em, itemsep=2pt, topsep=2pt, parsep=0pt]
    \item We identify the compound impact of output verbosity, sparse rewards, and tool noise on agentic search, which explains why monolithic architectures struggle with training stability (Figure~\ref{fig:challenges} and Section~\ref{sec:training_stability}).
    \item We introduce M-ASK, a collaborative framework that decouples search behavior from knowledge management, utilizing dense, turn-specific rewards to enable joint optimization.
    \item Extensive evaluations on multi-hop QA benchmarks demonstrate that M-ASK significantly outperforms state-of-the-art baselines, delivering robust gains in accuracy and markedly improving convergence stability in multi-hop search scenarios.
\end{itemize}

\section{Related Work}

\subsection{From Iterative RAG to Agentic Search}
The paradigm of Retrieval-Augmented Generation (RAG) has evolved significantly from static ``retrieve-then-read'' pipelines~\cite{ma2023query,shi2023towards,ke2024bridging,shi-etal-2024-generate,shi2025direct} to dynamic systems. Early iterative approaches, such as IRCoT~\cite{trivedi2023interleaving}, Self-RAG~\cite{asai2023self} and GEM~\cite{shi2025iterative}, introduced feedback loops where LLMs actively decide when to retrieve. Recent advancements have formalized this into \textbf{Agentic Search}, where models plan multi-step trajectories to solve open-ended problems using external tools.
Notably, emerging \textit{Deep Research} agents, such as DeepResearcher~\cite{zheng2025deepresearcher}, Search-o1~\cite{li2025search}, Search-r1~\cite{jin2025search}, and the R1-Searcher series~\cite{song2025r1, song2025r1++}, integrate retrieval directly into the reasoning chains of LLM to tackle long-horizon QA tasks.

\subsection{Context Management and Optimization}

To overcome static context limitations, recent works explore dynamic optimization via uncertainty-based filtering~\cite{jimenez2024hipporag,ji2025memory} or explicit memory agents~\cite{yu2025memagent,yan2025memory}. More aggressively, DeepNote~\cite{wang2024deepnote} and MemSearcher~\cite{yuan2025memsearcher} treat context as an evolving state. However, DeepNote does not support joint end-to-end optimization, while MemSearcher’s monolithic design leads to coarse credit assignment due to sparse, trajectory-level rewards. M-ASK addresses these pitfalls by \textit{decoupling} search and knowledge management into two specialized agents. Through multi-agent reinforcement learning with turn-specific dense rewards, our approach achieves precise credit assignment.

\subsection{Multi-Agent Systems for Information Retrieval} 
To overcome the limitations of monolithic agents, multi-agent systems (MAS) commonly decompose complex tasks into sub-tasks and assign them to agents with specialized roles.  General frameworks, such as MetaGPT~\cite{hong2023metagpt} and AutoGen~\cite{wu2024autogen}, exemplify this paradigm by enabling structured role specialization and coordination among multiple agents. 
Within the information retrieval domain, MindSearch~\cite{chen2024mindsearch} utilizes a graph-based planner alongside parallel web searchers for query decomposition, while MMOA-RAG~\cite{chen2025improving} allocates specialized agents for query rewriting, document selection, and answer generation. Similarly, MAO-ARAG~\cite{chen2025mao} focuses on optimizing a planner agent to balance retrieval effectiveness with efficiency. Despite their potential, most search-centric MAS rely heavily on prompt engineering or standard supervised fine-tuning (SFT), often neglecting the optimization of \textit{interaction dynamics}. Although MMOA-RAG incorporates MAPPO~\cite{yu2022surprising}, aims to optimize a shared global reward (e.g., the F1 score of the final answer). We argue that such sparse global feedback is suboptimal for multi-turn collaboration, as it exacerbates the \textit{credit assignment problem}, failing to accurately evaluate intermediate steps. M-ASK addresses this limitation by introducing distinct, dense reward functions for Search and Knowledge agents, ensuring that each role is optimized for its specific contribution to the collective goal.

\section{Method}
\label{sec:method}

We propose \textbf{M-ASK}, a multi-agent framework decoupling search planning from information integration. In this section, we first formulate the problem as a \textit{Sequential Decentralized Partially Observable Markov Decision Process}~\cite{bernstein2002complexity, oliehoek2016concise}. Second, we detail the specifications of the specialized agents. Finally, we present the framework's execution flow and the joint optimization process utilizing a turn-level reward mechanism.

\subsection{Problem Formulation}

We formulate the multi-hop QA task as a \textit{Sequential Decentralized Partially Observable Markov Decision Process}.
Unlike standard formulations where agents act simultaneously, our framework, M-ASK, operates in a turn-based manner.
Specifically, at each discrete time step $t$, only one designated agent, denoted as $\pi_{\text{active}}$, is activated. This agent receives a partial observation of the environment and executes an action to advance the search process.
To enable coordination across time steps, agents communicate indirectly by reading from and writing to a shared, structured knowledge state.



\paragraph{Structured Knowledge State}
Agents communicate via a shared, structured \textit{Knowledge State} $\mathcal{K}$. We define $\mathcal{K}_t$ at time step $t$ as:
\begin{equation}
\label{eq:state}
\mathcal{K}_t = \left\{
    \begin{aligned}
        & \text{``question"}: q, \\
        & \text{``thinking\_trajectory"}: \mathcal{T}_t, \\
        & \text{``predicted\_answer"}: a_t
    \end{aligned}
\right\}
\end{equation}
Let $\mathcal{T}_t = [ \tau_1, \tau_2, \dots, \tau_m ]$ represent the evolving reasoning chain, where each element $\tau_i = \langle q_{\text{sub}}^{(i)}, a_{\text{sub}}^{(i)} \rangle$ is a tuple consisting of a \textbf{sub-query} and its associated \textbf{sub-answer} (evidence).
This trajectory $\mathcal{T}_t$ explicitly records the step-by-step multi-hop inference process, serving as the logical derivation path required to deduce the final answer $a_t$ from the original question $q$.
Specifically, the \textbf{Planning Agent} initializes both the reasoning chain $\mathcal{T}_0$ and the answer $a_0$. As the multi-round search progresses, $\mathcal{T}_t$ is dynamically modified, and consequently, the answer $a_t$ is iteratively updated based on the evolving trajectory. The specific implementation details are discussed in the following section.
For a concrete instantiation of $\mathcal{K}_t$ in a multi-hop scenario, please refer to Table~\ref{tab:state_example} in Appendix~\ref{app:state_example}.

\begin{figure*}[t]
  \centering
  \includegraphics[width=1.0\textwidth]{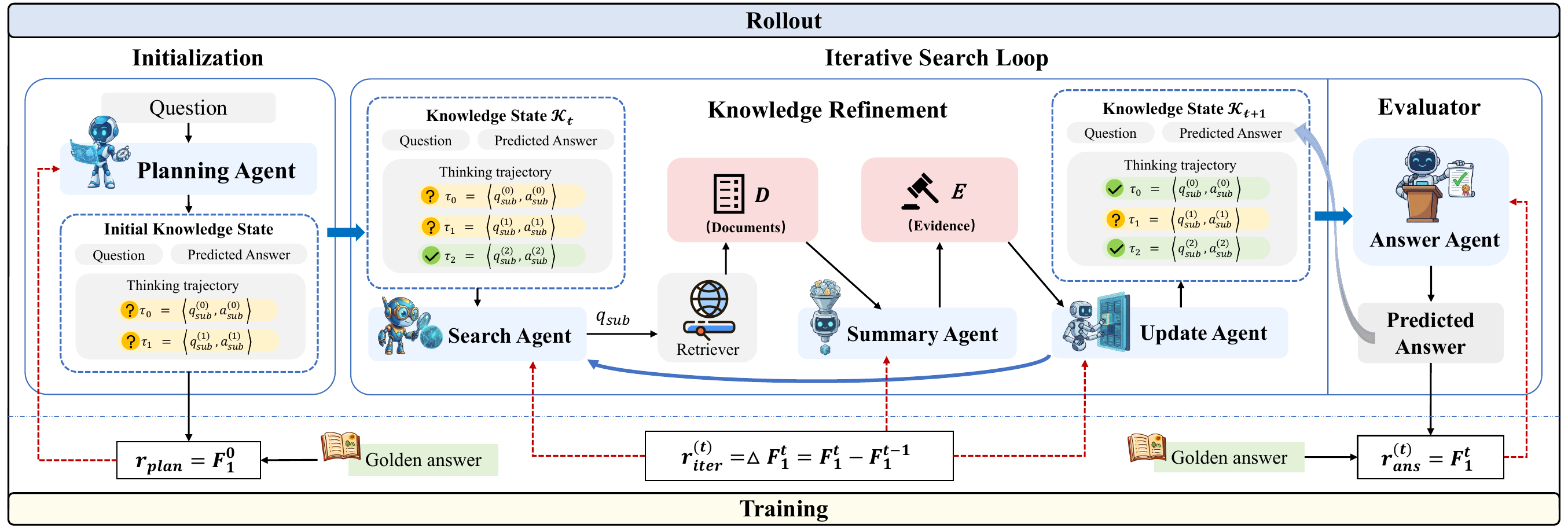}
  \caption{Overview of the M-ASK framework. (1) \textbf{Rollout}: The Planning Agent initializes the state $\mathcal{K}_0$, followed by an iterative loop where Search and Knowledge Management Agents refine the trajectory. Crucially, the Answer Agent updates the prediction after each turn. (2) \textbf{Training}: A hybrid reward mechanism assigns absolute scores ($F_1^0$ and $F_1^t$) to the Planning and Answer Agents, respectively, while the collaborative agents (Search, Summary, Update) share the marginal improvement ($\Delta F_1^t$) to incentivize step-wise refinement.}
  \label{fig:framework}
\end{figure*}

The framework shown in Figure \ref{fig:framework} orchestrates these agents into a cohesive workflow involving initialization, iterative refinement, and optimization (see Algorithm \ref{alg:mask_training_shared} in Appendix \ref{sec:appendix_training_algo} for the complete pseudo-code). 
To better understand how this workflow operates in practice, we first present the functional roles of the agents, followed by a detailed explanation of the inference and training processes.

\subsection{M-ASK Architecture}
\label{sec:framework}

M-ASK employs a team of specialized agents, categorized by their functional roles.

\subsubsection{Search Behavior Agents (SBA)}

\paragraph{Planning Agent ($A_{\text{plan}}$)}
This agent functions as the system \textit{initializer}. Given a user query $q$ as input, $A_{\text{plan}}$ leverages parametric memory to generate an initial reasoning trajectory $\mathcal{T}_0$ and a preliminary answer $a_0$, encapsulating them into the initial state $\mathcal{K}_0 \leftarrow \pi_{\text{plan}}(q)$.

\paragraph{Search Agent ($A_{\text{search}}$)}
Acting as the \textit{navigator}, this agent iteratively evaluates the sufficiency of the thinking trajectory $\mathcal{T}_t$ with respect to the query $q$. Its primary role is to decide between exploration and termination. Specifically, taking the question and trajectory as input, the policy outputs an action $Act \leftarrow \pi_{\text{search}}(q, \mathcal{T}_t)$, where $Act \in \{q'_{\text{sub}}, \texttt{<end>}\}$. If expanding knowledge is necessary, it generates a specific sub-query $q'_{\text{sub}}$; otherwise, it outputs \texttt{<end>} to terminate the search loop.

\paragraph{Answer Agent ($A_{\text{ans}}$)}
Operating as the \textit{solver}, this agent is responsible for generating the final prediction. Conditioned on the original question $q$ and the accumulated reasoning trajectory $\mathcal{T}_t$ retrieved from the knowledge state $\mathcal{K}$, it synthesizes a coherent final answer $a$, formally defined as $a \leftarrow \pi_{\text{ans}}(q, \mathcal{T}_t)$.

\subsubsection{Knowledge Management Agents (KMA)}

\paragraph{Summary Agent ($A_{\text{sum}}$)}
This agent functions as a \textit{filter} to distill key information. Given a sub-query $q'_{\text{sub}}$ and a set of retrieved documents $D$, it extracts the pertinent evidence $E$ while actively discarding irrelevant noise. The process is formalized as $E \leftarrow \pi_{\text{sum}}(q'_{\text{sub}}, D)$.

\paragraph{Update Agent ($A_{\text{upd}}$)}
Transcending the role of a passive logger, this agent acts as a \textit{dynamic state refiner}. Its primary objective is to maintain a high-density knowledge state $\mathcal{K}$ by judiciously deciding between refining existing information or appending new findings.
Given the current state $\mathcal{K}_t$, a sub-query $q'_{\text{sub}}$, and the retrieved evidence $E$, the agent outputs a discrete operation $op$ to evolve the trajectory. The action space is designed to balance information growth and precision:
(1) \texttt{<Update>$\tau_i$</Update>} (\textbf{In-Place Refinement}): Targeting an existing step $\tau_i$, this action overwrites previous hallucinations or vague information with precise evidence.
(2) \texttt{<Add>$\tau_{\text{new}}$</Add>} (\textbf{Expansion}): This action appends a new reasoning step only when the evidence introduces a distinct, necessary logical hop.
Formally, the state transition is defined as $op, \mathcal{K}_{t+1} \leftarrow \pi_{\text{upd}}(\mathcal{K}_t, q'_{\text{sub}}, E)$.

We provide a structured summary of the functional roles and specifications for each agent in Table~\ref{tab:appendix_agent_details} (Appendix~\ref{Detailed Agent Specifications}), offering a clear view of their distinct mechanisms.

\subsubsection{Inference Workflow}
The inference process operates as follows:
\begin{enumerate}[leftmargin=*, itemsep=2pt, topsep=2pt, parsep=0pt]
    \item \textbf{Initialization:} $A_{\text{plan}}$ generates a "cold start" state $\mathcal{K}_0$.
    \item \textbf{Iteration:} The system enters a loop controlled by $A_{\text{search}}$.
    \begin{itemize}[leftmargin=*, itemsep=2pt, topsep=2pt, parsep=0pt]
        \item At step $t$, $A_{\text{search}}$ evaluates $\mathcal{K}_t$.
        \item If $A_{\text{search}}$ generates a query, the workflow proceeds to $A_{\text{sum}}$ (filtering) and $A_{\text{upd}}$ (state update), producing $\mathcal{K}_{t+1}$.
    \end{itemize}
    \item \textbf{Termination:} The loop terminates when $A_{\text{search}}$ outputs \texttt{<end>} or a maximum step limit is reached. Finally, $A_{\text{ans}}$ is triggered to synthesize the final answer from the latest knowledge state $\mathcal{K}$.
\end{enumerate}

For a microscopic view of this workflow, including how agents dynamically correct hallucinations and refine the knowledge state, please refer to the detailed case study in Table \ref{tab:case_study_full} (Appendix \ref{app:case_study_full}).

\subsection{M-ASK Training via Turn-Level Rewards}
We employ Independent PPO ~\cite{schulman2017proximal} for optimization. To align the SBA and KMA groups despite their different functional roles, we design a hybrid reward mechanism that differentiates between \textit{state utilization} and \textit{state refinement}.

\paragraph{State-Based Reward (Absolute F1)}
For agents responsible for generating solution outputs ($A_{\text{plan}}$ and $A_{\text{ans}}$), the objective is to maximize the absolute quality of the current state.
\begin{equation}
    r_{\text{plan}} = \text{F}_1(a_0, y), \quad r_{\text{ans}}^{(t)} = \text{F}_1(a_t, y)
\label{f1_reward}
\end{equation}
Here, $a_t$ is the answer synthesized from $\mathcal{K}_t$ and $y$ is the ground truth answer. Note that during training, $A_{\text{ans}}$ acts as an evaluator at every step $t$.

\paragraph{Transition-Based Reward (Shared Incremental Gain)}

Crucially, the iterative phase requires tight collaboration between the Search Behavior Agent ($A_{\text{search}}$) and the Knowledge Management Agents ($A_{\text{sum}}, A_{\text{upd}}$). Although they belong to different functional groups, their actions are co-dependent: effective search requires precise state updates, and useful updates depend on accurate retrieval.

%
To enforce this \textbf{local cooperation}, we assign a \textbf{shared incremental reward} to all agents active in the loop. 
It is important to distinguish the execution frequency of the Answer Agent between phases. 
During \textbf{inference}, the Answer Agent is triggered only \textit{once} after the search terminates to generate the final output. 
However, during \textbf{training}, it assumes an additional role as an \textit{intermediate evaluator}. It is invoked at every turn $t$ to synthesize a temporary answer $a_t$ based on the current state $\mathcal{K}_t$. 
This mechanism allows us to utilize the answer score $r_{\text{ans}}^{(t)} = \text{F}_1(a_t, y)$ (defined in Eq.~\ref{f1_reward}) to measure the immediate answer quality. The iteration reward is then defined as the marginal improvement over the previous step:
\begin{equation}
    r_{\text{iter}}^{(t)} = r_{\text{ans}}^{(t)} - r_{\text{ans}}^{(t-1)} = \text{F}_1(a_t, y) - \text{F}_1(a_{t-1}, y)
\label{iter_reward}
\end{equation}
By sharing the identical $r_{\text{iter}}^{(t)}$ signal, the Search Agent ($A_{\text{search}}$), Summary Agent ($A_{\text{sum}}$), and Update Agent ($A_{\text{upd}}$) are jointly incentivized to maximize the marginal information gain of each turn. This mechanism binds them into a cooperative sub-team, where the Search Agent learns to fetch necessary information and the Knowledge Agents learn to distill it efficiently.
If $A_{\text{search}}$ outputs \texttt{<end>}, it receives a reward of $0$, ensuring the team only terminates the collaboration when further search yields no positive gain.

\paragraph{Optimization Objective (Parameter Sharing)}

To enhance sample efficiency and enable knowledge transfer across different reasoning phases, we employ a \textbf{Parameter-Shared}~\footnote{We provide a detailed justification for adopting this parameter-shared architecture in Appendix~\ref{app:parameter_sharing}.} strategy. All functional agents are instantiated from a unified LLM $\pi_\theta$, distinguished solely by role-specific system instructions $I_{\text{role}}$.

Consequently, the optimization objective aggregates the experiences from all roles. Let $u_t$ denote the action taken given observation $o_t$. The shared policy $\pi_\theta$ is updated to maximize:
\begin{equation}
\small
\begin{aligned}
    \mathcal{L}(\theta) = \hat{\mathbb{E}}_{t} \Big[ \min \Big( & \rho_t \hat{A}_t, \\
    & \text{clip}(\rho_t, 1-\epsilon, 1+\epsilon) \hat{A}_t \Big) \Big]
\end{aligned}
\end{equation}
where $\rho_t = \frac{\pi_{\theta}(u_t|o_t)}{\pi_{\theta_{old}}(u_t|o_t)}$ is the probability ratio.
Simultaneously, the shared Critic $V_\phi$ minimizes the unified value loss:
\begin{equation}
    \mathcal{L}(\phi) = \hat{\mathbb{E}}_{t} \left[ \| V_{\phi}(o_t) - R_t \|^2 \right]
\end{equation}
Here, $R_t = \sum_{k=0}^{T-t-1} \gamma^k r_{t+k}$ represents the cumulative discounted return, where $\gamma$ is the discount factor and $r$ denotes the role-specific rewards defined earlier.

\begin{table*}[!ht]
\centering
\small
\resizebox{\textwidth}{!}{
\begin{tabular}{l|cccc|ccccc|c}
\toprule
\textbf{Method} & \multicolumn{4}{c|}{\textbf{Single-hop QA}} & \multicolumn{5}{c|}{\textbf{Multi-hop QA}} & \textbf{Avg} \\
& \textbf{NQ} & \textbf{PopQA} & \textbf{AmbigQA} & \textbf{Avg} & \textbf{HotpotQA} & \textbf{2Wiki} & \textbf{Musique} & \textbf{Bam.} & \textbf{Avg} & \textbf{All} \\
\midrule
\multicolumn{11}{l}{\textit{Standard Baselines}} \\
LLM w/o RAG & 17.53 & 14.93 & 23.53 & 18.66 & 17.76 & 22.58 & 8.58 & 17.14 & 16.52 & 17.44 \\
Vanilla RAG & 40.60 & 42.74 & 56.20 & 46.51 & 28.33 & 25.91 & 25.20 & 25.28 & 26.18 & 34.89 \\
\midrule
\multicolumn{11}{l}{\textit{RL-Based (Static Modular Workflow)}} \\
RRR~\cite{ma2023query} & 54.60 & \textbf{50.46} & 65.41 & 56.82 & 46.21 & 41.52 & 18.27 & 36.59 & 35.65 & 44.72 \\
BGM~\cite{ke2024bridging} & 54.21 & 49.51 & 65.97 & 56.56 & 46.85 & 37.79 & 17.55 & 37.38 & 34.89 & 44.18 \\
MMOA-RAG~\cite{chen2025improving} & \underline{55.44} & \underline{50.21} & \underline{68.02} & \underline{57.89} & 49.21 & 41.66 & 17.26 & 37.20 & 36.33 & 45.57 \\
\midrule
\multicolumn{11}{l}{\textit{Agentic Search (Adaptive Workflow)}} \\
Adaptive RAG~\cite{jeong2024adaptive} & 36.52 & 35.59 & 45.32 & 39.14 & 42.38 & 39.62 & \underline{25.48} & 34.85 & 35.58 & 37.11 \\
MAO-ARAG~\cite{chen2025mao} & 36.82 & 41.85 & 47.03 & 41.90 & 46.65 & \underline{43.96} & 22.38 & \textbf{49.84} & \underline{40.85} & 41.30 \\
DeepNote~\cite{wang2024deepnote} & 54.03 & 49.80 & 67.57 & 57.13 & \underline{52.49} & 36.22 & 22.17 & \underline{45.22} & 39.03 & \underline{46.79} \\
Search-r1~\cite{jin2025search} & 52.57 & 46.98 & 65.25 & 54.93 & 46.87 & 39.03 & 17.97 & 38.69 & 35.64 & 43.91 \\
\midrule
\textbf{M-ASK (Ours)} & \textbf{57.40} & 50.10 & \textbf{68.33} & \textbf{58.61} & \textbf{58.31} & \textbf{46.12} & \textbf{26.20} & 44.18 & \textbf{43.70} & \textbf{50.09} \\
\textit{Impv. vs Best} & \textcolor{blue}{+1.96} & \textcolor{gray}{-0.36} & \textcolor{blue}{+0.31} & \textcolor{blue}{+0.72} & \textcolor{blue}{+5.82} & \textcolor{blue}{+2.16} & \textcolor{blue}{+0.72} & \textcolor{gray}{-5.66} & \textcolor{blue}{+2.85} & \textcolor{blue}{+3.30} \\
\bottomrule
\end{tabular}
}
\caption{Main performance comparison (F1 Score) on single-hop and multi-hop QA benchmarks. The best results are \textbf{bolded} and the second best are \underline{underlined}. "Impv. vs Best" denotes the performance gain (\textcolor{blue}{blue}) or drop (\textcolor{gray}{gray}) of M-ASK compared to the best performing baseline in each column.}
\label{tab:main_results_avg}
\end{table*}

\section{Experiments}
\label{sec:experiments}

To validate the efficacy of M-ASK, we conduct comprehensive experiments to answer the following research questions: \textbf{RQ1}: Does M-ASK outperform existing monolithic and multi-agent frameworks? \textbf{RQ2}: How does M-ASK compare to the monolithic baseline (Search-r1) in terms of training stability? \textbf{RQ3}: Is it beneficial to jointly model and optimize Knowledge Management and Search Behavior? \textbf{RQ4}: Do turn-specific dense rewards provide better credit assignment than global outcome-based rewards?

\subsection{Experimental Setup}

\paragraph{Datasets}
We evaluate our framework on a diverse set of open-domain QA benchmarks categorized by reasoning complexity. For \textbf{Single-hop QA}, we use Natural Questions (NQ)~\cite{kwiatkowski2019natural}, PopQA~\cite{mallen2022not}, and AmbigQA~\cite{min2020ambigqa} to test factual retrieval accuracy. For \textbf{Multi-hop QA}, to evaluate complex reasoning and trajectory planning, we employ HotpotQA~\cite{yang2018hotpotqa}, 2WikiMultiHopQA~\cite{ho2020constructing}, Musique~\cite{trivedi2022musique}, and Bamboogle~\cite{press2022measuring}.

\paragraph{Baselines}
We evaluate M-ASK against three categories of methods: \textbf{Standard Baselines}, \textbf{RL-Based Static Workflows}, and \textbf{Adaptive Agentic Search}. Detailed implementations and settings for these baselines are provided in Appendix \ref{app:baseline_details}.

\paragraph{Implementation Details}
Our implementation is developed based on the official \texttt{verl} repository\footnote{\url{https://github.com/volcengine/verl}}. 
For all experiments, we use \textbf{Qwen2.5-7B-Instruct}~\cite{team2024qwen2} as the backbone LLM. We utilize the English Wikipedia as our retrieval corpus, indexed via \textbf{E5}~\cite{wang2022text} for dense retrieval.
We report the standard \textbf{F1 Score} as the evaluation metric.

\subsection{Main Results Analysis (RQ1)}
\label{sec:main_results_analysis}

To answer \textbf{RQ1}, we evaluate M-ASK against competitive baselines across seven datasets. As shown in Table~\ref{tab:main_results_avg}, M-ASK achieves the highest average F1 score (\textbf{50.09}), outperforming both monolithic and multi-agent frameworks.

\paragraph{Dominance on Complex QA and Generalization.}
M-ASK demonstrates superior capability on complex reasoning tasks, particularly on \textbf{HotpotQA}, where it surpasses the best baseline by a substantial margin of \textbf{+5.82}. This validates that our collaborative framework is far more effective at handling complex multi-hop queries than standard methods. Furthermore, M-ASK generalizes robustly to unseen out-of-domain datasets (e.g., \textbf{2Wiki}, \textbf{Musique}), consistently exceeding other Agentic baselines. This suggests M-ASK learns abstract, transferable strategies for query decomposition and filtering rather than merely overfitting to training patterns.

\paragraph{Analysis of Bamboogle Performance.}
A notable exception is observed on \textbf{Bamboogle}, where M-ASK trails MAO-ARAG. We attribute this to two primary factors. First, the test set of Bamboogle is significantly smaller (only 125 queries) than other benchmarks, introducing potential statistical variance. Second, MAO-ARAG utilizes frozen pre-trained answer generators, which may answer factoid questions based on their existing parametric knowledge, whereas M-ASK is optimized to ground answers strictly in retrieved evidence. Nevertheless, M-ASK remains highly competitive, trailing the second-best method (DeepNote) by only $\sim$1.0 point while outperforming all other baselines, reaffirming the robustness of our policy despite these factors.

\paragraph{Comparison with Monolithic and Modular Frameworks.}
A clear performance hierarchy is observed, further answering RQ1 regarding framework efficacy. The \textbf{monolithic} agent, Search-r1, lags significantly (43.91) due to \textit{unconstrained context growth} and accumulated search noise. This structural limitation is addressed by \textbf{modular} frameworks like DeepNote (46.79), which achieves the second-best results by decoupling knowledge management. However, M-ASK further outperforms DeepNote (+3.30). Unlike DeepNote's disjoint modules, M-ASK employs \textit{end-to-end joint optimization}, ensuring that the Search and Knowledge Management agents are collaboratively updated to maximize the final reasoning reward.

\subsection{Training Stability and Convergence (RQ2)}
\label{sec:training_stability}

\begin{figure*}[h]
    \centering
    \begin{minipage}{0.31\textwidth}
        \centering
        \includegraphics[width=\textwidth]{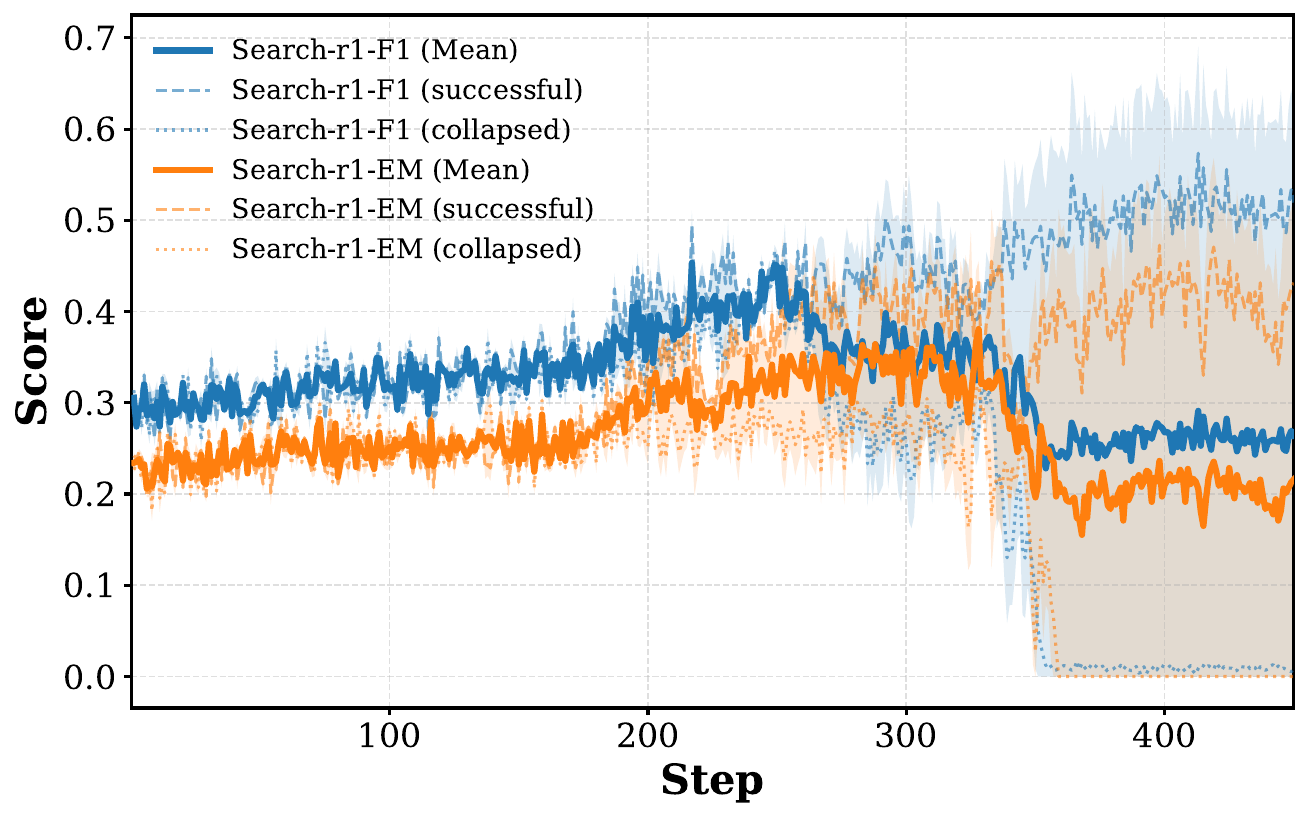}
        \caption*{(a) Search-r1 Training Dynamics}
    \end{minipage}
    \hfill
    \begin{minipage}{0.33\textwidth}
        \centering
        \includegraphics[width=\textwidth]{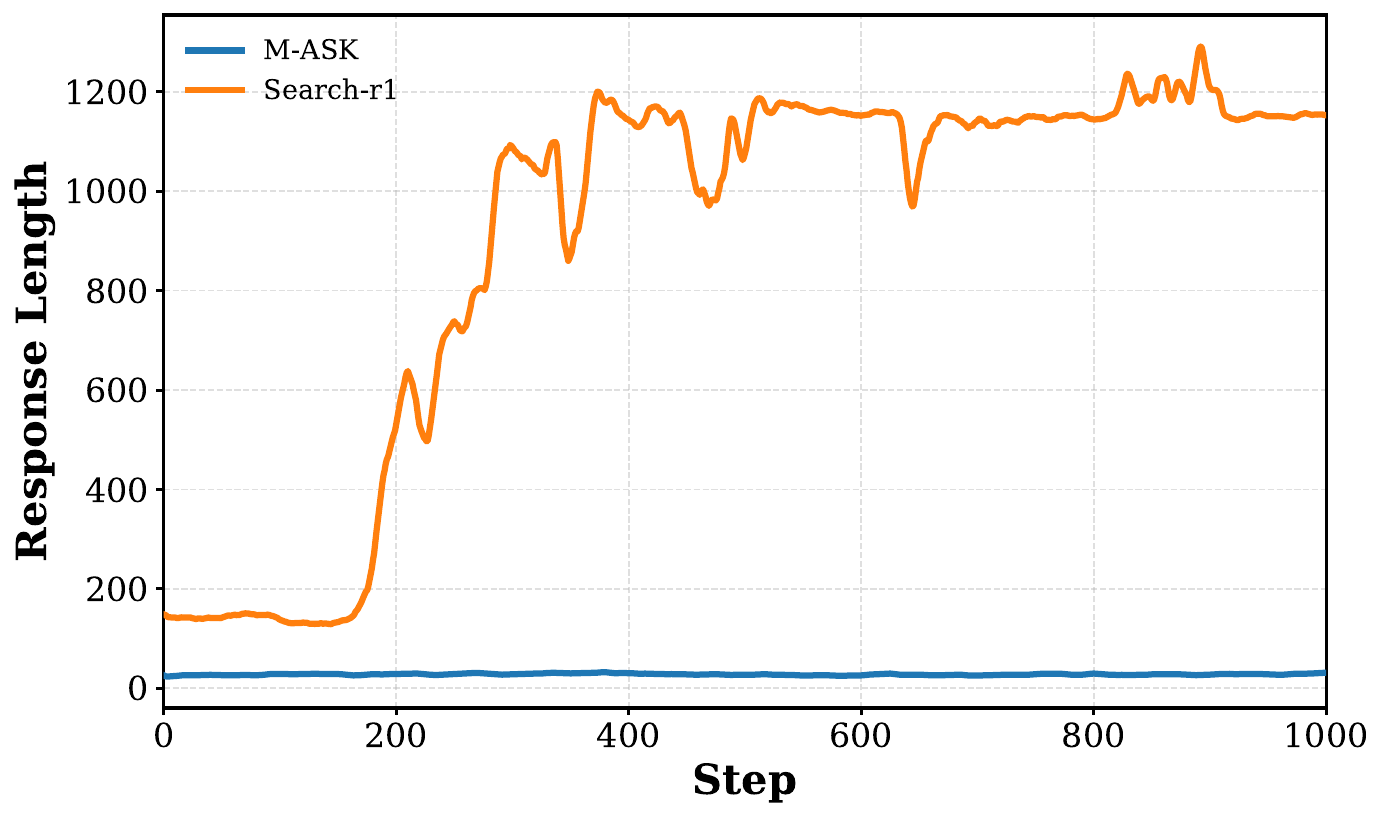}
        \caption*{(b) Average Response Length}
    \end{minipage}
    \hfill
    \begin{minipage}{0.32\textwidth}
        \centering
        \includegraphics[width=\textwidth]{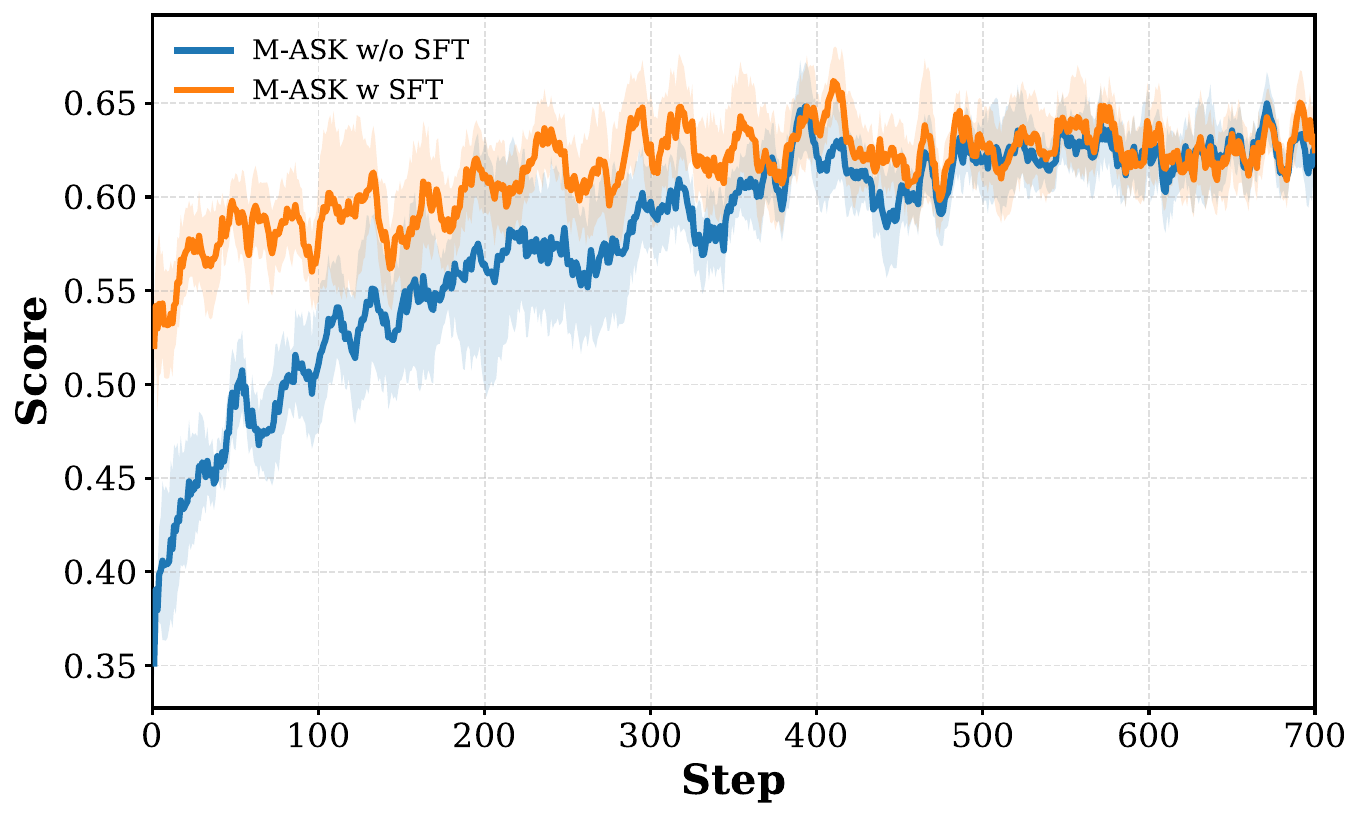}
        \caption*{(c) M-ASK Training Dynamics}
    \end{minipage}
    \hfill
    \caption{Training curves on HotpotQA. In \textbf{(a)} and \textbf{(c)}, solid lines represent the mean across multiple runs, while shaded regions indicate the variance. \textbf{(a)} Search-r1 exhibits high instability and frequent mode collapse. \textbf{(b)} Evolution of average response length; M-ASK remains concise while Search-r1 suffers from context bloating. \textbf{(c)} M-ASK demonstrates stable convergence.}
    \label{fig:training_curves}
\end{figure*}

To investigate the training dynamics and stability of our framework compared to monolithic approaches, we visualize the training curves on the HotpotQA dataset in Figure~\ref{fig:training_curves}. Furthermore, to rigorously quantify robustness, we conducted 10 independent training runs for both methods and recorded the rate of "model collapse"—defined as the performance score dropping to near zero and failing to recover. These statistics are summarized in Table~\ref{tab:collapse_rate}.

\begin{table}[h]
\centering
\small
\setlength{\tabcolsep}{2pt}
\begin{tabular}{l|ccc}
\toprule
\multirow{2}{*}{\textbf{Method}} & \multicolumn{3}{c}{\textbf{Collapse Rate (over 10 runs)}} \\
& \textbf{@ 200 Steps} & \textbf{@ 500 Steps} & \textbf{@ 1000 Steps} \\
\midrule
Search-r1 & 1/10 (10\%) & 7/10 (70\%) & 9/10 (90\%) \\
\textbf{M-ASK (Ours)} & \textbf{0/10 (0\%)} & \textbf{0/10 (0\%)} & \textbf{0/10 (0\%)} \\
\bottomrule
\end{tabular}
\caption{Training stability analysis comparing the rate of model collapse (performance degrading to $\approx 0$) at different training stages across 10 independent runs.}
\label{tab:collapse_rate}
\end{table}

\paragraph{Catastrophic Collapse in Monolithic Agents.}
As shown in Table~\ref{tab:collapse_rate} and Figure~\ref{fig:training_curves}(a), the monolithic Search-r1 agent exhibits extreme volatility. Consistent with the findings in~\cite{deng2025grpo}, we observe that Search-r1 is prone to collapse during the mid-to-late stages of training. Specifically, while only 10\% of runs failed at 200 steps, the collapse rate escalated to \textbf{90\%} as training progressed to 1000 steps.
We attribute this to the interaction between long horizons and sparse rewards. As the agent explores, it inevitably encounters retrieval noise. In a monolithic architecture, this noise accumulates in the context without intermediate correction. When the optimization algorithm attempts to assign credit based solely on the final outcome, the noisy gradients frequently push the policy into degenerate states from which it cannot recover.

\paragraph{Analysis of Response Length and Mechanism.}
The key to M-ASK's superior stability (0\% collapse rate) is elucidated in Figure~\ref{fig:training_curves}(b). While the monolithic Search-r1 suffers from "context bloating"—rapidly converging to over 1000 tokens—M-ASK maintains a remarkably low and stable average output length. This conciseness is not merely a cosmetic difference but a direct indicator of three structural advantages that stabilize training: (1) \textbf{Simplified Functionality:} Decoupling ensures atomic tasks with short generation horizons, significantly reducing optimization complexity compared to monolithic models; (2) \textbf{Precise Credit Assignment:} Short outputs allow turn-specific dense rewards to provide immediate, high-confidence feedback, unlike global rewards that dilute over long trajectories; and (3) \textbf{Effective Noise Filtering:} Knowledge Management Agents actively prune noise to prevent error accumulation in the knowledge state.


Collectively, these factors eliminate the gradient variance caused by noisy, long contexts, and sparse reward. This structural robustness enables M-ASK to achieve the stable, monotonic convergence illustrated in Figure~\ref{fig:training_curves}(c), even in the absence of supervised warm-starting.

\begin{table*}[ht]
\centering
\scriptsize 
\setlength{\tabcolsep}{1.2pt} 
\resizebox{\textwidth}{!}{
\begin{tabular}{l ccccccccc c cccccccccc}
\toprule
\multirow{3}{*}{\textbf{Method}} & 
\multicolumn{8}{c}{\textbf{Single-hop Datasets}} & &
\multicolumn{10}{c}{\textbf{Multi-hop Datasets}} \\
\cmidrule(lr){2-9} \cmidrule(lr){11-20}

& \multicolumn{2}{c}{NQ} & \multicolumn{2}{c}{PopQA} & \multicolumn{2}{c}{AmbigQA} & \multicolumn{2}{c}{\textbf{Avg.}} & &
\multicolumn{2}{c}{HotpotQA} & \multicolumn{2}{c}{2Wiki} & \multicolumn{2}{c}{Musique} & \multicolumn{2}{c}{Bam.} & \multicolumn{2}{c}{\textbf{Avg.}} \\
\cmidrule(lr){2-3} \cmidrule(lr){4-5} \cmidrule(lr){6-7} \cmidrule(lr){8-9} 
\cmidrule(lr){11-12} \cmidrule(lr){13-14} \cmidrule(lr){15-16} \cmidrule(lr){17-18} \cmidrule(lr){19-20}

& F1 & $\Delta$ & F1 & $\Delta$ & F1 & $\Delta$ & \textbf{F1} & \textbf{$\Delta$} & &
F1 & $\Delta$ & F1 & $\Delta$ & F1 & $\Delta$ & F1 & $\Delta$ & \textbf{F1} & \textbf{$\Delta$} \\
\midrule

\textbf{M-ASK (Ours)} & 
57.40 & -- & 50.10 & -- & 68.33 & -- & \textbf{58.61} & -- & &
58.31 & -- & 46.12 & -- & 26.20 & -- & 44.18 & -- & \textbf{43.70} & -- \\
\midrule
\multicolumn{20}{l}{\textit{Ablation Studies}} \\

w/o KMA & 
54.93 & \color{gray}{-2.47} & 48.36 & \color{gray}{-1.74} & 64.79 & \color{gray}{-3.54} & \textbf{56.03} & \textbf{\color{gray}{-2.58}} & &
55.06 & \color{gray}{-3.25} & 46.58 & \color{blue}{+0.46} & 19.39 & \color{gray}{-6.81} & 42.33 & \color{gray}{-1.85} & \textbf{40.84} & \textbf{\color{gray}{-2.86}} \\

w/o T-L Reward & 
54.95 & \color{gray}{-2.45} & 47.12 & \color{gray}{-2.98} & 62.04 & \color{gray}{-6.29} & \textbf{54.70} & \textbf{\color{gray}{-3.91}} & &
42.57 & \color{gray}{-15.74} & 33.27 & \color{gray}{-12.85} & 11.95 & \color{gray}{-14.25} & 26.10 & \color{gray}{-18.08} & \textbf{28.47} & \textbf{\color{gray}{-15.23}} \\

\bottomrule
\end{tabular}
}
\caption{Ablation study. \textbf{Avg.} columns show the aggregated performance on Single-hop (3 datasets) and Multi-hop (4 datasets) benchmarks. $\Delta$ represents the performance drop (gray) or gain (\textcolor{blue}{blue}) compared to the full model.}
\label{tab:ablation_grouped}
\end{table*}

\subsection{Ablation Studies (RQ3 \& RQ4)}
\label{sec:ablation}

To validate the contribution of individual components in M-ASK, we conduct ablation studies and observing the impact across different task complexities. The results are summarized in Table~\ref{tab:ablation_grouped}.

\subsubsection{Impact of Collaborative Knowledge Management (RQ3)}
To answer whether explicitly modeling knowledge management is beneficial, we evaluate the variant \textbf{w/o KMA}. In this setting, we remove the \textit{Summary} and \textit{Update} agents; the \textit{Search Behavior Agent} interacts directly with the raw search engine output and appends full documents to the context. 
This setup simulates the unconstrained information flow of a standard monolithic interaction. However, unlike the fully monolithic \textit{Search-r1}, this variant retains the underlying multi-agent paradigm and independent optimization methods, thereby strictly isolating the impact of the missing knowledge management module.

\paragraph{Analysis.}
The results show a consistent performance degradation across both single-hop (avg. $\Delta -2.58\%$) and multi-hop (avg. $\Delta -2.86\%$) benchmarks. This confirms that without the active filtering provided by the KMA group, the search agent is overwhelmed by noise in the retrieval results, leading to hallucinations or distracted reasoning.
We note a slight anomaly on \textbf{2WikiMultiHopQA} ($+0.46\%$), where the unfiltered model performs marginally better. 
However, this is outweighed by the severe drops in more complex datasets like \textbf{Musique} ($-6.81\%$) and \textbf{HotpotQA} ($-3.25\%$). Overall, explicit knowledge management is crucial for stabilizing the reasoning trajectory against retrieval noise.

\subsubsection{Efficacy of Turn-Specific Dense Rewards (RQ4)}
To assess the necessity of our reward shaping mechanism, we evaluate \textbf{w/o T-L Reward}. In this variant, we disable the turn-level incremental rewards (Eq.\ref{f1_reward} and Eq.\ref{iter_reward}) and instead optimize all agents using only the final outcome reward (Global F1) at the end of the episode, similar to the strategy used in MMOA-RAG~\cite{chen2025improving}.

\paragraph{Analysis.}
This ablation reveals a critical insight into the optimization dynamics of multi-turn search. \textbf{Single-hop Resilience:} On single-hop datasets, the performance drop is moderate (avg. $\Delta -3.91\%$) because the short trajectories typical of these tasks make the final reward sufficient for credit assignment. \textbf{Multi-hop Collapse:} In stark contrast, performance collapses on multi-hop datasets (avg. $\Delta -15.23\%$), with the score on \textbf{HotpotQA} plummeting by over 15 points.

This disparity highlights the \textbf{turn-level credit assignment problem}. In frameworks like MMOA-RAG, all agents across different turns share an identical global outcome reward. This ambiguity obscures the specific contribution of individual steps, making it difficult to discern which action actually drove the success or failure. Consequently, agents struggle to optimize intermediate sub-goals in long-horizon tasks. In contrast, M-ASK's turn-specific $\Delta \text{F1}$ reward provides immediate, dense feedback, enabling agents to accurately lock in optimal policies for each reasoning hop.

\section{Conclusion}
\label{sec:conclusion}

In this paper, we addressed the structural instability of monolithic agentic search caused by unconstrained contexts, sparse rewards, and search noise. We proposed \textbf{M-ASK}, a multi-agent framework that decouples search planning from knowledge management, enabling synergistic collaboration via turn-specific dense rewards.

Empirical evaluations across seven benchmarks demonstrate that M-ASK consistently outperforms state-of-the-art baselines, particularly in complex multi-hop scenarios. Crucially, our analysis reveals that decoupling these roles significantly enhances training stability compared to end-to-end RL approaches. These findings suggest that explicit role specialization and intermediate supervision are critical for scaling agentic search to more open-ended and noisy real-world environments. Future work will explore extending M-ASK to heterogeneous model architectures and broader tool-use scenarios beyond information retrieval.


\section*{Limitations}
\label{sec:limitations}

Despite the effectiveness of M-ASK, several limitations remain. 
First, the collaborative multi-agent workflow inherently increases the frequency of LLM invocations compared to single-pass approaches, as each reasoning step requires discrete inference calls for search, summarization, and update modules. This sequential interaction pattern inevitably raises computational costs and may pose challenges for latency-sensitive applications.
Second, our evaluation is currently confined to textual QA tasks; the framework's generalizability to other complex domains, such as code generation or multimodal reasoning, remains to be explored. 
Finally, the efficacy of our parameter-sharing strategy relies on the inherent capacity of the backbone LLM to handle diverse role instructions, and performance on smaller-scale architectures warrants further investigation.

\section*{Ethics Statement}
\label{sec:ethics}

This work builds upon large language models to facilitate multi-agent collaboration for textual question answering. 
All datasets used in our experiments are publicly available and do not contain any personally identifiable information. 
Nevertheless, outputs generated by LLMs may reflect biases present in their pretraining data or produce incorrect information. 
Users and practitioners should exercise caution when applying the proposed framework to real-world scenarios, especially in domains where misinformation or biased content could cause harm. 
We do not foresee additional ethical risks beyond those commonly associated with contemporary language model research.

\bibliography{acl}

@article{asai2023self,
  title={Self-rag: Learning to retrieve, generate, and critique through self-reflection},
  author={Asai, Akari and Wu, Zeqiu and Wang, Yizhong and Sil, Avirup and Hajishirzi, Hannaneh},
  journal={arXiv preprint arXiv:2310.11511},
  year={2023}
}

@article{ma2023query,
  title={Query rewriting for retrieval-augmented large language models},
  author={Ma, Xinbei and Gong, Yeyun and He, Pengcheng and Zhao, Hai and Duan, Nan},
  journal={arXiv preprint arXiv:2305.14283},
  year={2023}
}

@article{ke2024bridging,
  title={Bridging the preference gap between retrievers and llms},
  author={Ke, Zixuan and Kong, Weize and Li, Cheng and Zhang, Mingyang and Mei, Qiaozhu and Bendersky, Michael},
  journal={arXiv preprint arXiv:2401.06954},
  year={2024}
}

@article{schulman2017proximal,
  title={Proximal policy optimization algorithms},
  author={Schulman, John and Wolski, Filip and Dhariwal, Prafulla and Radford, Alec and Klimov, Oleg},
  journal={arXiv preprint arXiv:1707.06347},
  year={2017}
}

@article{yu2022surprising,
  title={The surprising effectiveness of ppo in cooperative multi-agent games},
  author={Yu, Chao and Velu, Akash and Vinitsky, Eugene and Gao, Jiaxuan and Wang, Yu and Bayen, Alexandre and Wu, Yi},
  journal={Advances in Neural Information Processing Systems},
  volume={35},
  pages={24611--24624},
  year={2022}
}

@article{yang2018hotpotqa,
  title={HotpotQA: A dataset for diverse, explainable multi-hop question answering},
  author={Yang, Zhilin and Qi, Peng and Zhang, Saizheng and Bengio, Yoshua and Cohen, William W and Salakhutdinov, Ruslan and Manning, Christopher D},
  journal={arXiv preprint arXiv:1809.09600},
  year={2018}
}

@article{ho2020constructing,
  title={Constructing a multi-hop QA dataset for comprehensive evaluation of reasoning steps},
  author={Ho, Xanh and Nguyen, Anh-Khoa Duong and Sugawara, Saku and Aizawa, Akiko},
  journal={arXiv preprint arXiv:2011.01060},
  year={2020}
}

@article{min2020ambigqa,
  title={AmbigQA: Answering ambiguous open-domain questions},
  author={Min, Sewon and Michael, Julian and Hajishirzi, Hannaneh and Zettlemoyer, Luke},
  journal={arXiv preprint arXiv:2004.10645},
  year={2020}
}

@article{jin2025search,
  title={Search-r1: Training llms to reason and leverage search engines with reinforcement learning},
  author={Jin, Bowen and Zeng, Hansi and Yue, Zhenrui and Yoon, Jinsung and Arik, Sercan and Wang, Dong and Zamani, Hamed and Han, Jiawei},
  journal={arXiv preprint arXiv:2503.09516},
  year={2025}
}

@article{song2025r1,
  title={R1-Searcher: Incentivizing the Search Capability in LLMs via Reinforcement Learning},
  author={Song, Huatong and Jiang, Jinhao and Min, Yingqian and Chen, Jie and Chen, Zhipeng and Zhao, Wayne Xin and Fang, Lei and Wen, Ji-Rong},
  journal={arXiv preprint arXiv:2503.05592},
  year={2025}
}

@article{wang2022text,
  title={Text embeddings by weakly-supervised contrastive pre-training},
  author={Wang, Liang and Yang, Nan and Huang, Xiaolong and Jiao, Binxing and Yang, Linjun and Jiang, Daxin and Majumder, Rangan and Wei, Furu},
  journal={arXiv preprint arXiv:2212.03533},
  year={2022}
}

@article{chen2025improving,
  title={Improving Retrieval-Augmented Generation through Multi-Agent Reinforcement Learning},
  author={Chen, Yiqun and Yan, Lingyong and Sun, Weiwei and Ma, Xinyu and Zhang, Yi and Wang, Shuaiqiang and Yin, Dawei and Yang, Yiming and Mao, Jiaxin},
  journal={arXiv preprint arXiv:2501.15228},
  year={2025}
}

@article{kwiatkowski2019natural,
  title={Natural questions: a benchmark for question answering research},
  author={Kwiatkowski, Tom and Palomaki, Jennimaria and Redfield, Olivia and Collins, Michael and Parikh, Ankur and Alberti, Chris and Epstein, Danielle and Polosukhin, Illia and Devlin, Jacob and Lee, Kenton and others},
  journal={Transactions of the Association for Computational Linguistics},
  volume={7},
  pages={453--466},
  year={2019},
  publisher={MIT Press One Rogers Street, Cambridge, MA 02142-1209, USA journals-info~…}
}

@article{mallen2022not,
  title={When not to trust language models: Investigating effectiveness and limitations of parametric and non-parametric memories},
  author={Mallen, Alex and Asai, Akari and Zhong, Victor and Das, Rajarshi and Hajishirzi, Hannaneh and Khashabi, Daniel},
  journal={arXiv preprint arXiv:2212.10511},
  volume={7},
  year={2022}
}

@article{trivedi2022musique,
  title={MuSiQue: Multihop Questions via Single-hop Question Composition},
  author={Trivedi, Harsh and Balasubramanian, Niranjan and Khot, Tushar and Sabharwal, Ashish},
  journal={Transactions of the Association for Computational Linguistics},
  volume={10},
  pages={539--554},
  year={2022},
  publisher={MIT Press One Broadway, 12th Floor, Cambridge, Massachusetts 02142, USA~…}
}

@article{press2022measuring,
  title={Measuring and narrowing the compositionality gap in language models},
  author={Press, Ofir and Zhang, Muru and Min, Sewon and Schmidt, Ludwig and Smith, Noah A and Lewis, Mike},
  journal={arXiv preprint arXiv:2210.03350},
  year={2022}
}

@article{team2024qwen2,
  title={Qwen2 technical report},
  author={Team, Qwen},
  journal={arXiv preprint arXiv:2412.15115},
  year={2024}
}

@article{hurst2024gpt,
  title={Gpt-4o system card},
  author={Hurst, Aaron and Lerer, Adam and Goucher, Adam P and Perelman, Adam and Ramesh, Aditya and Clark, Aidan and Ostrow, AJ and Welihinda, Akila and Hayes, Alan and Radford, Alec and others},
  journal={arXiv preprint arXiv:2410.21276},
  year={2024}
}

@article{li2025search,
  title={Search-o1: Agentic search-enhanced large reasoning models},
  author={Li, Xiaoxi and Dong, Guanting and Jin, Jiajie and Zhang, Yuyao and Zhou, Yujia and Zhu, Yutao and Zhang, Peitian and Dou, Zhicheng},
  journal={arXiv preprint arXiv:2501.05366},
  year={2025}
}

@inproceedings{trivedi2023interleaving,
  title={Interleaving retrieval with chain-of-thought reasoning for knowledge-intensive multi-step questions},
  author={Trivedi, Harsh and Balasubramanian, Niranjan and Khot, Tushar and Sabharwal, Ashish},
  booktitle={Proceedings of the 61st annual meeting of the association for computational linguistics (volume 1: long papers)},
  pages={10014--10037},
  year={2023}
}

@article{zheng2025deepresearcher,
  title={Deepresearcher: Scaling deep research via reinforcement learning in real-world environments},
  author={Zheng, Yuxiang and Fu, Dayuan and Hu, Xiangkun and Cai, Xiaojie and Ye, Lyumanshan and Lu, Pengrui and Liu, Pengfei},
  journal={arXiv preprint arXiv:2504.03160},
  year={2025}
}

@inproceedings{hong2023metagpt,
  title={MetaGPT: Meta programming for a multi-agent collaborative framework},
  author={Hong, Sirui and Zhuge, Mingchen and Chen, Jonathan and Zheng, Xiawu and Cheng, Yuheng and Wang, Jinlin and Zhang, Ceyao and Wang, Zili and Yau, Steven Ka Shing and Lin, Zijuan and others},
  booktitle={The Twelfth International Conference on Learning Representations},
  year={2023}
}

@inproceedings{wu2024autogen,
  title={Autogen: Enabling next-gen LLM applications via multi-agent conversations},
  author={Wu, Qingyun and Bansal, Gagan and Zhang, Jieyu and Wu, Yiran and Li, Beibin and Zhu, Erkang and Jiang, Li and Zhang, Xiaoyun and Zhang, Shaokun and Liu, Jiale and others},
  booktitle={First Conference on Language Modeling},
  year={2024}
}

@article{chen2024mindsearch,
  title={Mindsearch: Mimicking human minds elicits deep ai searcher},
  author={Chen, Zehui and Liu, Kuikun and Wang, Qiuchen and Liu, Jiangning and Zhang, Wenwei and Chen, Kai and Zhao, Feng},
  journal={arXiv preprint arXiv:2407.20183},
  year={2024}
}

@article{chen2025mao,
  title={Mao-arag: Multi-agent orchestration for adaptive retrieval-augmented generation},
  author={Chen, Yiqun and Zhang, Erhan and Yan, Lingyong and Wang, Shuaiqiang and Huang, Jizhou and Yin, Dawei and Mao, Jiaxin},
  journal={arXiv preprint arXiv:2508.01005},
  year={2025}
}

@article{ji2025memory,
  title={Memory-Aware and Uncertainty-Guided Retrieval for Multi-Hop Question Answering},
  author={Ji, Yuelyu and Meng, Rui and Li, Zhuochun and He, Daqing},
  journal={arXiv preprint arXiv:2503.23095},
  year={2025}
}

@article{yan2025memory,
  title={Memory-r1: Enhancing large language model agents to manage and utilize memories via reinforcement learning},
  author={Yan, Sikuan and Yang, Xiufeng and Huang, Zuchao and Nie, Ercong and Ding, Zifeng and Li, Zonggen and Ma, Xiaowen and Kersting, Kristian and Pan, Jeff Z and Sch{\"u}tze, Hinrich and others},
  journal={arXiv preprint arXiv:2508.19828},
  year={2025}
}

@article{yu2025memagent,
  title={MemAgent: Reshaping Long-Context LLM with Multi-Conv RL-based Memory Agent},
  author={Yu, Hongli and Chen, Tinghong and Feng, Jiangtao and Chen, Jiangjie and Dai, Weinan and Yu, Qiying and Zhang, Ya-Qin and Ma, Wei-Ying and Liu, Jingjing and Wang, Mingxuan and others},
  journal={arXiv preprint arXiv:2507.02259},
  year={2025}
}

@article{song2025r1++,
  title={R1-Searcher++: Incentivizing the Dynamic Knowledge Acquisition of LLMs via Reinforcement Learning},
  author={Song, Huatong and Jiang, Jinhao and Tian, Wenqing and Chen, Zhipeng and Wu, Yuhuan and Zhao, Jiahao and Min, Yingqian and Zhao, Wayne Xin and Fang, Lei and Wen, Ji-Rong},
  journal={arXiv preprint arXiv:2505.17005},
  year={2025}
}

@article{rashid2020monotonic,
  title={Monotonic value function factorisation for deep multi-agent reinforcement learning},
  author={Rashid, Tabish and Samvelyan, Mikayel and De Witt, Christian Schroeder and Farquhar, Gregory and Foerster, Jakob and Whiteson, Shimon},
  journal={Journal of Machine Learning Research},
  volume={21},
  number={178},
  pages={1--51},
  year={2020}
}

@article{lowe2017multi,
  title={Multi-agent actor-critic for mixed cooperative-competitive environments},
  author={Lowe, Ryan and Wu, Yi I and Tamar, Aviv and Harb, Jean and Pieter Abbeel, OpenAI and Mordatch, Igor},
  journal={Advances in neural information processing systems},
  volume={30},
  year={2017}
}

@article{chen2022ptde,
  title={PTDE: Personalized training with distilled execution for multi-agent reinforcement learning},
  author={Chen, Yiqun and Mao, Hangyu and Mao, Jiaxin and Wu, Shiguang and Zhang, Tianle and Zhang, Bin and Yang, Wei and Chang, Hongxing},
  journal={arXiv preprint arXiv:2210.08872},
  year={2022}
}

@article{deng2025grpo,
  title={On GRPO Collapse in Search-R1: The Lazy Likelihood-Displacement Death Spiral},
  author={Deng, Wenlong and Li, Yushu and Gong, Boying and Ren, Yi and Thrampoulidis, Christos and Li, Xiaoxiao},
  journal={arXiv preprint arXiv:2512.04220},
  year={2025}
}

@article{wang2024deepnote,
  title={DeepNote: Note-Centric Deep Retrieval-Augmented Generation},
  author={Wang, Ruobing and Zhao, Qingfei and Yan, Yukun and Zha, Daren and Chen, Yuxuan and Yu, Shi and Liu, Zhenghao and Wang, Yixuan and Wang, Shuo and Han, Xu and others},
  journal={arXiv preprint arXiv:2410.08821},
  year={2024}
}

@article{jimenez2024hipporag,
  title={Hipporag: Neurobiologically inspired long-term memory for large language models},
  author={Jimenez Gutierrez, Bernal and Shu, Yiheng and Gu, Yu and Yasunaga, Michihiro and Su, Yu},
  journal={Advances in Neural Information Processing Systems},
  volume={37},
  pages={59532--59569},
  year={2024}
}

@article{yuan2025memsearcher,
  title={MemSearcher: Training LLMs to Reason, Search and Manage Memory via End-to-End Reinforcement Learning},
  author={Yuan, Qianhao and Lou, Jie and Li, Zichao and Chen, Jiawei and Lu, Yaojie and Lin, Hongyu and Sun, Le and Zhang, Debing and Han, Xianpei},
  journal={arXiv preprint arXiv:2511.02805},
  year={2025}
}

@article{jeong2024adaptive,
  title={Adaptive-rag: Learning to adapt retrieval-augmented large language models through question complexity},
  author={Jeong, Soyeong and Baek, Jinheon and Cho, Sukmin and Hwang, Sung Ju and Park, Jong C},
  journal={arXiv preprint arXiv:2403.14403},
  year={2024}
}

@article{rafailov2023direct,
  title={Direct preference optimization: Your language model is secretly a reward model},
  author={Rafailov, Rafael and Sharma, Archit and Mitchell, Eric and Manning, Christopher D and Ermon, Stefano and Finn, Chelsea},
  journal={Advances in neural information processing systems},
  volume={36},
  pages={53728--53741},
  year={2023}
}

@misc{shi2025deepresearchsystematicsurvey,
      title={Deep Research: A Systematic Survey}, 
      author={Zhengliang Shi and Yiqun Chen and Haitao Li and Weiwei Sun and Shiyu Ni and Yougang Lyu and Run-Ze Fan and Bowen Jin and Yixuan Weng and Minjun Zhu and Qiujie Xie and Xinyu Guo and Qu Yang and Jiayi Wu and Jujia Zhao and Xiaqiang Tang and Xinbei Ma and Cunxiang Wang and Jiaxin Mao and Qingyao Ai and Jen-Tse Huang and Wenxuan Wang and Yue Zhang and Yiming Yang and Zhaopeng Tu and Zhaochun Ren},
      year={2025},
      eprint={2512.02038},
      archivePrefix={arXiv},
      primaryClass={cs.CL},
      url={https://arxiv.org/abs/2512.02038}, 
}

@article{bernstein2002complexity,
  title={The complexity of decentralized control of Markov decision processes},
  author={Bernstein, Daniel S and Givan, Robert and Immerman, Neil and Zilberstein, Shlomo},
  journal={Mathematics of operations research},
  volume={27},
  number={4},
  pages={819--840},
  year={2002},
  publisher={INFORMS}
}

@book{oliehoek2016concise,
  title={A concise introduction to decentralized POMDPs},
  author={Oliehoek, Frans A and Amato, Christopher and others},
  volume={1},
  year={2016},
  publisher={Springer}
}

@article{shi2025iterative,
  title={Iterative self-incentivization empowers large language models as agentic searchers},
  author={Shi, Zhengliang and Yan, Lingyong and Yin, Dawei and Verberne, Suzan and de Rijke, Maarten and Ren, Zhaochun},
  journal={arXiv preprint arXiv:2505.20128},
  year={2025}
}

@inproceedings{shi-etal-2024-generate,
    title = "Generate-then-Ground in Retrieval-Augmented Generation for Multi-hop Question Answering",
    author = "Shi, Zhengliang  and
      Zhang, Shuo  and
      Sun, Weiwei  and
      Gao, Shen  and
      Ren, Pengjie  and
      Chen, Zhumin  and
      Ren, Zhaochun",
    editor = "Ku, Lun-Wei  and
      Martins, Andre  and
      Srikumar, Vivek",
    booktitle = "Proceedings of the 62nd Annual Meeting of the Association for Computational Linguistics (Volume 1: Long Papers)",
    month = aug,
    year = "2024",
    address = "Bangkok, Thailand",
    publisher = "Association for Computational Linguistics",
    url = "https://aclanthology.org/2024.acl-long.397/",
    doi = "10.18653/v1/2024.acl-long.397",
    pages = "7339--7353",
}

@article{shi2025direct,
  title={Direct retrieval-augmented optimization: Synergizing knowledge selection and language models},
  author={Shi, Zhengliang and Yan, Lingyong and Sun, Weiwei and Feng, Yue and Ren, Pengjie and Ma, Xinyu and Wang, Shuaiqiang and Yin, Dawei and de Rijke, Maarten and Ren, Zhaochun},
  journal={arXiv preprint arXiv:2505.03075},
  year={2025}
}

@inproceedings{shi2023towards,
  title={Towards a unified framework for reference retrieval and related work generation},
  author={Shi, Zhengliang and Gao, Shen and Zhang, Zhen and Chen, Xiuying and Chen, Zhumin and Ren, Pengjie and Ren, Zhaochun},
  booktitle={Findings of the Association for Computational Linguistics: EMNLP 2023},
  pages={5785--5799},
  year={2023}
}

\clearpage
\appendix

\section{Rationale for Parameter Sharing Strategy}
\label{app:parameter_sharing}

In the M-ASK framework, we employ a parameter-sharing strategy where a single Large Language Model (LLM) backbone $\pi_\theta$ serves as the underlying policy for all functional agents (i.e., $A_{\text{plan}}, A_{\text{search}}, A_{\text{sum}}, A_{\text{upd}}, A_{\text{ans}}$), distinguished solely by role-specific system instructions. We adopt this design based on three critical considerations:

\paragraph{1. Theoretical Foundation in MARL}
Parameter sharing is a well-established and effective paradigm in Multi-Agent Reinforcement Learning (MARL)~\cite{rashid2020monotonic,lowe2017multi,yu2022surprising,chen2022ptde}. In cooperative settings, sharing parameters allows agents to learn a unified representation of the state space, which often leads to faster convergence and improved training stability compared to maintaining independent policies. By optimizing a single set of parameters on the aggregated experiences of all roles, the model can efficiently generalize across the diverse phases of the search and reasoning process.

\paragraph{2. Computational and Storage Efficiency}
Unlike traditional RL agents based on small Multi-Layer Perceptrons (MLPs), agents in our framework are initialized with LLMs containing billions of parameters. Maintaining independent policy networks for five distinct agents would result in a linear increase in memory consumption ($\mathcal{O}(N)$), rendering the training process computationally prohibitive and difficult to deploy.
Parameter sharing reduces the storage requirement to $\mathcal{O}(1)$, significantly lowering the barrier for training and inference. This efficiency is paramount for complex RAG systems like M-ASK, allowing us to allocate resources toward longer context windows rather than redundant model weights.

\paragraph{3. Inherent Multi-Task Capability of LLMs}
LLMs inherently possess strong multi-task capabilities, enabling them to perform distinct tasks based on contextual instructions (prompts) without modifying their internal weights.
In M-ASK, different agents (e.g., the \textit{Search Agent} deciding on queries vs. the \textit{Summary Agent} extracting evidence) share fundamental reasoning competencies, such as reading comprehension and logical deduction. Parameter sharing leverages this synergy: skills learned in the \textit{summarization} task can implicitly enhance the \textit{answer generation} capability via shared representations. By conditioning the shared $\pi_\theta$ on role-specific prompts $I_{role}$, we effectively project the model's general capabilities into specific functional subspaces, achieving role specialization without architectural redundancy.

\section{Implementation Details of Baselines}
\label{app:baseline_details}

We compare M-ASK against three categories of methods:
\begin{itemize}
    \item \textbf{Standard Baselines:} \textit{LLM w/o RAG} (closed-book), \textit{Vanilla RAG} (standard retrieve-then-generate).
    \item \textbf{RL-Based (Static Modular Workflow):} \textit{RRR}~\cite{ma2023query} (RL-based query reformulation), and \textit{BGM}~\cite{ke2024bridging} (RL-based document selection), and \textit{MMOA-RAG}~\cite{chen2025improving} (multi-agent RL).
    \item \textbf{Agentic Search (Adaptive Workflow):} \textit{Adaptive RAG}~\cite{jeong2024adaptive} (adaptive workflow), \textit{MAO-ARAG}~\cite{chen2025mao} (planner-executors optimization), \textit{DeepNote}~\cite{wang2024deepnote} (knowledge- management) and \textit{Search-r1}~\cite{jin2025search} (monolithic agent optimized via RL reasoning training).
\end{itemize}

To ensure a fair comparison, all baselines and our M-ASK method are unified under the same experimental setting. We utilize \texttt{Qwen2.5-7B-Instruct} as the backbone model. Specifically, for components within the baselines that do not require training, we employ the pre-trained version of \texttt{Qwen2.5-7B-Instruct}. For components requiring training, we fine-tune them based on the \texttt{Qwen2.5-7B-Instruct} initialization. Detailed implementation notes for each baseline are provided below:

\paragraph{Standard Baselines}
\begin{itemize}
    \item \textbf{LLM w/o RAG:} This represents the closed-book setting where the LLM directly generates answers based on its internal parametric knowledge without accessing external corpora.
    \item \textbf{Vanilla RAG:} A standard retrieve-then-generate pipeline. It retrieves the top-5 relevant documents based on the query and concatenates them with the prompt to generate the answer using the pre-trained LLM.
\end{itemize}

\paragraph{Static Modular Workflow (RL-Based)}
\begin{itemize}
    \item \textbf{RRR}~\cite{ma2023query}: This method employs the PPO algorithm to end-to-end train a query rewriter module. In our reproduction, to prevent the performance bottleneck potentially caused by a frozen generator, we also fine-tuned the answer generation module, ensuring it is well-trained alongside the rewriter.
    \item \textbf{BGM}~\cite{ke2024bridging}: BGM uses PPO to train a document selection module. Although the original implementation may utilize a frozen generator, we fine-tuned the answer generation module in our reproduction to align with the robust setting of RRR and maximize overall performance.
    \item \textbf{MMOA-RAG}~\cite{chen2025improving}: This framework utilizes Multi-Agent PPO (MAPPO)~\cite{yu2022surprising} to train three distinct agents—a query rewriter, a document selector, and an answer generator—guided by a shared final reward mechanism. We adopted \texttt{Qwen2.5-7B-Instruct} as the backbone model for this multi-agent framework, strictly following the original logic while maintaining consistency with our unified setting.
\end{itemize}

\paragraph{Agentic Search (Adaptive Workflow)}
\begin{itemize}
    \item \textbf{Adaptive RAG}~\cite{jeong2024adaptive}: This method trains a classifier to dynamically determine which of three distinct retrieval workflows (Directly Answer, Simple RAG, or Iterative RAG) should handle a given query. In our implementation, we fine-tuned \texttt{Qwen2.5-7B-Instruct} to serve as this classifier, while utilizing the pre-trained version of \texttt{Qwen2.5-7B-Instruct} for executing the subsequent static workflows.
    \item \textbf{MAO-ARAG}~\cite{chen2025mao}: MAO-ARAG features a hierarchical multi-agent architecture composed of a planner and multiple executors, sharing a modular structure similar to our M-ASK. The original work utilizes PPO with a shared final reward to train specifically the planner agent. Consistent with our unified setting, the planner module is a fine-tuned \texttt{Qwen2.5-7B-Instruct}, while the executor modules employ the pre-trained version of the same model.
    \item \textbf{DeepNote}~\cite{wang2024deepnote}: DeepNote focuses on optimizing knowledge management within QA tasks through Direct Preference Optimization (DPO)~\cite{rafailov2023direct}. In our reproduction, we upgraded the data generation model from \texttt{gpt-4o-mini}~\cite{hurst2024gpt} (used in the original paper) to \texttt{gpt-4o}~\cite{hurst2024gpt} to construct higher-quality DPO training data.
    \item \textbf{Search-r1}~\cite{jin2025search}: Unlike modular designs, Search-r1 adopts a monolithic architecture where multi-turn search and reasoning processes occur within a single response generation. It is optimized via RL using outcome-based rewards. For this baseline, we obtained the experimental results using the code provided in the author's official open-source repository\footnote{\url{https://github.com/PeterGriffinJin/Search-R1}}.
\end{itemize}

\section{Prompt for Different Agents}

\subsection{Prompt for Planning Agent}

The following is the full system prompt used for decomposing queries. The \texttt{[EXAMPLE\_PROMPT]} placeholder represents the few-shot examples injected at runtime.

\begin{promptbox}[title=System Prompt]
You are a knowledgeable reasoning assistant. Your task is to decompose a given question into multiple sub-questions based purely on your own knowledge (do not use external tools), and answer each sub-question based on your own knowledge. If you do not know the answer, write 'unkown' exactly.\\

Output all results strictly in the following format for easy parsing:
\textless{}q1\textgreater{}sub-question text\textless{}/q1\textgreater{}\\
\textless{}a1\textgreater{}answer text\textless{}/a1\textgreater{}\\
\textless{}q2\textgreater{}sub-question text\textless{}/q2\textgreater{}\\
\textless{}a2\textgreater{}answer text\textless{}/a2\textgreater{}\\

Continue numbering sequentially until done. At the end, output your final conclusion answer to the original question in the tag \textless{}predicted\_answer\textgreater{}...\textless{}/predicted\_answer\textgreater{}, and inside this tag follow the style of:
[EXAMPLE\_PROMPT]
\end{promptbox}

\begin{promptbox}[title=User Input]
Query: \{query\}\\
\end{promptbox}

\subsection{Prompt for Search Agent}

\begin{promptbox}[title=System Prompt]
You are a search query generation assistant.\\
Your input contains:\\
- The original question\\
- A thinking\_trajectory (list of sub-questions and their sub-answers)\\
- A search\_history (list of past search queries and their summaries)\\

Note:\\
- The search\_history represents queries YOU have already searched in previous turns.\\
- You MUST avoid generating queries that are identical to or semantically similar to any query in search\_history.\\
- Only generate a new query if it provides new, distinct information that has not yet been searched.\\

Your task:\\
1. If any sub-question in thinking\_trajectory has an answer 'unkown' or unclear, generate a specific, searchable query to find the missing information.\\
2. If you believe additional sub-questions are needed to answer the original question, generate a specific searchable query for that.\\
3. If no further search is needed, output \textless{}end\textgreater{}.\\

Output only one of the following two formats:\\
\textless{}search\textgreater{}Your query here\textless{}/search\textgreater{}\\
\textless{}end\textgreater{}\\

Do not include any extra text, explanations, or other tags.
\end{promptbox}

\begin{promptbox}[title=User Input]
Original question: \{question\}\\

Thinking trajectory:\\
\{trajectory\_text\}\\

Search history:\\
\{history\_text\}\\

Output only one of the following two formats:\\
\textless{}search\textgreater{}Your query here\textless{}/search\textgreater{}\\
\textless{}end\textgreater{}\\

Do not include any extra text, explanations, or other tags.
\end{promptbox}

\subsection{Prompt for Summary Agent}

\begin{promptbox}[title=System Prompt]
You are an evidence extraction assistant.\\
You will be given:\\
- A search query\\
- Several candidate documents related to that query\\

Your task:\\
1. From the provided documents, extract the key evidence that directly answers the query.\\
2. If no key evidence exists in the documents, output 'No useful information'.\\

Output exactly in one of the two formats:\\
\textless{}evidence\textgreater{}your key evidence here\textless{}/evidence\textgreater{}\\
\textless{}evidence\textgreater{}No useful information\textless{}/evidence\textgreater{}\\

Do not include explanations, reasoning steps, or any text outside the \textless{}evidence\textgreater{} tags.
\end{promptbox}

\begin{promptbox}[title=User Input]
Query: \{query\}\\
Candidate Documents:
\{docs\_text\}
\end{promptbox}

\subsection{Prompt for Update Agent}

\begin{promptbox}[title=System Prompt]
You are an update detection assistant.\\
You are given:\\
- An original question and its thinking\_trajectory (sub-questions and their answers)\\
- The predicted\_answer derived from thinking\_trajectory\\
- A new question and its evidence\\

Your task:\\
1. If the new question and evidence can replace one existing sub-question-answer pair in thinking\_trajectory, output \textless{}Update\textgreater{}ti\textless{}/Update\textgreater{} where ti is the key.\\
2. If the new question and evidence contains information not present in thinking\_trajectory, output \textless{}Add\textgreater{}t(n+1)\textless{}/Add\textgreater{} where n is current number of t's.\\
3. Output strictly one of these formats and no other text.
\end{promptbox}

\begin{promptbox}[title=User Input]
Original question: \{orig\_question\}\\
Thinking trajectory:
\{trajectory\_text\}\\
New query: \{new\_query\}\\
Evidence: \{new\_evidence\}
\end{promptbox}

\subsection{Prompt for Answer Agent}

\begin{promptbox}[title=System Prompt]
You are a final answer generation assistant.\\
You are given:\\
- An original question\\
- A thinking\_trajectory (list of sub-questions and their answers)\\

Your task:\\
1. Use the provided thinking\_trajectory's answers to determine the final answer to the original question.\\
2. You must only output in the following strict format as shown in [EXAMPLE\_PROMPT]\\

IMPORTANT: Output must be:\\
\textless{}predicted\_answer\textgreater{}your answer here\textless{}/predicted\_answer\textgreater{}\\
No extra text, no reasoning.
\end{promptbox}

\begin{promptbox}[title=User Input]
Question: \{question\}\\
Thinking trajectory: \{trajectory\_text\}
\end{promptbox}

\section{Detailed Agent Specifications}
\label{Detailed Agent Specifications}

In this section, we provide comprehensive specifications for the multi-agent architecture employed in M-ASK. Due to space constraints in the main text, we present the granular details of the Search Behavior Agents (SBA) and Knowledge Management Agents (KMA) here. Table~\ref{tab:appendix_agent_details} formally defines the specific roles, input-output interfaces, and action spaces for each agent, elucidating their operational logic within the iterative retrieval process.

\begin{table*}[t]
\centering
\renewcommand{\arraystretch}{1.35} 
\setlength{\tabcolsep}{6pt}        
\resizebox{\textwidth}{!}{%
\begin{tabular}{l|l|l|p{7.5cm}|p{5.5cm}}
\toprule
\textbf{Module} & \textbf{Agent} & \textbf{Role} & \textbf{Description \& Action Space} & \textbf{Input / Output Formulation} \\
\midrule

\multirow{10}{*}{\textbf{SBA}} 
& \textbf{Planning} 
& \textit{Initializer} 
& Leverages parametric memory to generate an initial reasoning trajectory $\mathcal{T}_0$ and a preliminary answer $a_0$, encapsulating them into the initial state.
& \textbf{In:} Query $q$ \newline
  \textbf{Out:} $\mathcal{K}_0 \leftarrow \pi_{\text{plan}}(q)$ \\
\cline{2-5}

& \textbf{Search} 
& \textit{Navigator} 
& Evaluates information sufficiency to decide between exploration and termination. \newline
1) Generate sub-query $q'_{\text{sub}}$ to expand knowledge. \newline
2) Output \texttt{<end>} to terminate the loop.
& \textbf{In:} Query $q$, Trajectory $\mathcal{T}_t$ \newline
  \textbf{Out:} $Act \leftarrow \pi_{\text{search}}(\mathcal{K}_t)$, \newline
  where $Act \in \{q'_{\text{sub}}, \texttt{<end>}\}$ \\
\cline{2-5}

& \textbf{Answer} 
& \textit{Solver} 
& Synthesizes the final answer prediction conditioned on the original question and the accumulated reasoning trajectory.
& \textbf{In:} Query $q$, Trajectory $\mathcal{T}_t$ \newline
  \textbf{Out:} $a \leftarrow \pi_{\text{ans}}(q, \mathcal{T}_t)$ \\
\midrule

\multirow{11}{*}{\textbf{KMA}} 
& \textbf{Summary} 
& \textit{Filter} 
& Distills pertinent evidence from retrieved documents while actively discarding irrelevant noise to ensure information density.
& \textbf{In:} Sub-query $q'_{\text{sub}}$, Documents $D$ \newline
  \textbf{Out:} $E \leftarrow \pi_{\text{sum}}(q'_{\text{sub}}, D)$ \\
\cline{2-5}

& \textbf{Update} 
& \textit{Dynamic Refiner} 
& Judiciously decides how to integrate evidence to evolve the trajectory: \newline
\textbullet\ \texttt{<Update>$\tau_i$</Update>} (\textbf{In-Place Refinement}): Overwrites hallucinations/vague steps. \newline
\textbullet\ \texttt{<Add>$\tau_{\text{new}}$</Add>} (\textbf{Expansion}): Appends necessary logical hops.
& \textbf{In:} State $\mathcal{K}_t$, sub-query $q'_{\text{sub}}$, Evidence $E$ \newline
  \textbf{Out:} $op, \mathcal{K}_{t+1} \leftarrow \pi_{\text{upd}}(\mathcal{K}_t, q'_{\text{sub}}, E)$ \\
\bottomrule
\end{tabular}%
}
\caption{Comprehensive specifications of the agents within the M-ASK framework, detailing their roles, functional mechanisms, and input/output formalizations.}
\label{tab:appendix_agent_details}
\end{table*}

\section{Example of Structured Knowledge State}
\label{app:state_example}

To facilitate a better understanding of the data structure defined in Eq.~\ref{eq:state}, we provide a concrete example of the Structured Knowledge State $\mathcal{K}_t$. 

Consider a multi-hop question $q$: \textit{``Who is the current CEO of the company that developed ChatGPT?''}. 
Table~\ref{tab:state_example} illustrates the state $\mathcal{K}_t$, where the agent has successfully decomposed the query and retrieved the necessary evidence chain.

\begin{table*}[h]
\centering
\small
\begin{tabular}{l p{0.8\linewidth}}
\toprule
\textbf{Key} & \textbf{Value (Content)} \\
\midrule
\texttt{"question"} ($q$) & Who is the current CEO of the company that developed ChatGPT? \\
\midrule
\texttt{"thinking\_trajectory"} ($\mathcal{T}_t$) & 
\begin{minipage}[t]{\linewidth}
    \textbf{($\tau_1$)}: \\
    \quad $\langle q_{sub}^{(1)} \rangle$: Which company developed ChatGPT? \\
    \quad $\langle a_{sub}^{(1)} \rangle$: OpenAI developed ChatGPT, an AI chatbot launched in November 2022. \\
    \textbf{($\tau_2$)}: \\
    \quad $\langle q_{sub}^{(2)} \rangle$: Who is the current CEO of OpenAI? \\
    \quad $\langle a_{sub}^{(2)} \rangle$: Sam Altman is the CEO of OpenAI. \\
\end{minipage} \\
\midrule
\texttt{"predicted\_answer"} ($a_t$) & Sam Altman \\
\bottomrule
\end{tabular}
\caption{A example of the Knowledge State $\mathcal{K}_t$. The trajectory $\mathcal{T}_t$ contains a sequence of sub-query and sub-answer pairs that logically support the final answer.}
\label{tab:state_example}
\end{table*}

\section{Joint Training Algorithm Details}
\label{sec:appendix_training_algo}

In this section, we provide the comprehensive pseudo-code for the M-ASK joint training process, complementing the methodology described in Section 3.3. Algorithm \ref{alg:mask_training_shared} details the execution flow of the \textbf{Parameter-Shared} strategy, where a unified policy $\pi_\theta$ is employed across all functional roles (Planning, Search, Summary, Update, and Answer).

The procedure consists of two distinct phases:
\begin{itemize}
    \item \textbf{Data Collection Phase:} Agents interact sequentially to solve the task. Notably, the Answer Agent serves as an intermediate evaluator at each step $t$, calculating the marginal performance gain $\Delta \text{F1}$ to provide dense, turn-specific supervision.
    \item \textbf{Unified Optimization Phase:} Experiences from all roles are stored in a mixed replay buffer $\mathcal{B}$. The shared model parameters $\theta$ and critic $\phi$ are then jointly updated via PPO, maximizing the objective function derived from these heterogeneous interaction trajectories.
\end{itemize}

\begin{algorithm*}[!h] 
\caption{M-ASK Joint Training (Parameter Shared)}
\label{alg:mask_training_shared}
\SetAlgoLined
\KwIn{Data $\mathcal{D}$, Unified Model $\pi_\theta$, Critic $V_\phi$, Role Instructions $I=\{I_{\text{plan}}, I_{\text{search}}, \dots\}$, Batch $B$}
\KwOut{Optimized Parameters $\theta^*, \phi^*$}

Initialize $\theta, \phi$; Initialize unified replay buffer $\mathcal{B}$\;

\For{iteration $i = 1, \dots, M$}{
    \tcp{1. Data Collection Phase}
    \While{$|\mathcal{B}| < B$}{
        Sample $(q, y) \sim \mathcal{D}$\;
        
        \tcp{Phase I: Initialization}
        $\mathcal{K}_0 \leftarrow \pi_{\theta}(q; I_{\text{plan}})$; $a_0 \leftarrow \mathcal{K}_0.\text{predicted\_answer}$\;
        $S_{prev} \leftarrow \text{F1}(a_0, y)$\;
        Add transition $(I_{\text{plan}}, q, \mathcal{K}_0, r=S_{prev})$ to $\mathcal{B}$\;
        
        \tcp{Phase II: Iterative Collaboration}
        $t \leftarrow 0$\;
        \While{$t < T_{max}$}{
            \tcp{Step 1: SBA Decision}
            $Action \leftarrow \pi_{\theta}(\mathcal{K}_t; I_{\text{search}})$ \tcp*[r]{Prompt as Searcher}
            
            \If{$Action == \texttt{<end>}$}{
                Add transition $(I_{\text{search}}, \mathcal{K}_t, \texttt{<end>}, r=0)$ to $\mathcal{B}$\;
                \textbf{break}\;
            }
            $q'_{sub} \leftarrow Action$\;

            \tcp{Step 2: KMA Execution}
            $D \leftarrow \text{SearchEngine}(q'_{sub})$\;
            $E \leftarrow \pi_{\theta}(q'_{sub}, D; I_{\text{sum}})$ \tcp*[r]{as Summarizer}
            $op, \mathcal{K}_{next} \leftarrow \pi_{\theta}(\mathcal{K}_t, q'_{sub}, E; I_{\text{upd}})$ \tcp*[r]{as Updater}
            
            \tcp{Step 3: Evaluation}
            $a_{next} \leftarrow \pi_{\theta}(\mathcal{K}_{next}; I_{\text{ans}})$ \tcp*[r]{as Answerer}
            $S_{curr} \leftarrow \text{F1}(a_{next}, y)$\;
            $\Delta \text{F1} \leftarrow S_{curr} - S_{prev}$\;
            
            \tcp{Step 4: Store Mixed Experiences}
            Add $(I_{\text{search}}, \mathcal{K}_t, q'_{sub}, r=\Delta \text{F1})$ to $\mathcal{B}$\;
            Add $(I_{\text{sum}}, q'_{sub}, D, E, r=\Delta \text{F1})$ to $\mathcal{B}$\;
            Add $(I_{\text{upd}}, \mathcal{K}_t, E, op, r=\Delta \text{F1})$ to $\mathcal{B}$\;
            Add $(I_{\text{ans}}, \mathcal{K}_{next}, a_{next}, r=S_{curr})$ to $\mathcal{B}$\;
            
            $\mathcal{K}_{t+1} \leftarrow \mathcal{K}_{next}$; $S_{prev} \leftarrow S_{curr}$; $t \leftarrow t + 1$\;
        }
    }
    
    \tcp{2. Unified Optimization Phase}
    Compute GAE on unified buffer $\mathcal{B}$ using shared Critic $V_\phi$\;
    \For{epoch $k = 1, \dots, K$}{
        Sample mixed mini-batches from $\mathcal{B}$\;
        Update $\theta, \phi$ via PPO maximizing joint objective $\mathcal{L}(\theta)$\;
    }
    Clear buffer $\mathcal{B}$\;
}
\end{algorithm*}

\section{Detailed Case Study: Structured Knowledge State Evolution}
\label{app:case_study_full}


The trajectory presented in Table \ref{tab:case_study_full} offers a microscopic view of how M-ASK addresses the structural limitations of monolithic agents. The execution process exhibits high rationality and ingenuity in three critical aspects:

\paragraph{1. Correction of Parametric Hallucinations via "Targeting".}
A pervasive challenge in agentic search is the "cold start" problem, where models must plan without initial observations. In the \textbf{Init} phase, the Planning Agent relies on its parametric memory and incorrectly predicts "Guadalajara."
Crucially, M-ASK does not treat this initial plan as ground truth but as a \textit{search target}. In \textbf{Turn 1}, the Update Agent identifies a conflict between the parametric "Guadalajara" and the retrieved non-parametric evidence ("Puebla"). Instead of attempting to reconcile the two or hedging, the agent prioritizes external evidence, executing an \texttt{<Update>} action to overwrite the hallucination. This demonstrates a robust \textbf{error-correction mechanism} that prevents initial errors from propagating through the reasoning chain.

\paragraph{2. Deep Verification of Temporal Constraints.}
The transition to \textbf{Turn 2} highlights the Search Agent's ability to handle complex constraints. The query contains a specific temporal condition ("1943"). While a naive retriever might stop after finding the entity "Puebla," the Search Agent actively verifies the historical context ("significant event... in 1943").
The subsequent \texttt{<Add>t2</Add>} action is highly rational: the agent recognizes that the "professionalization of the league" is distinct background information that supports the answer's validity. By appending this as a new step ($t_2$), the system constructs a logical evidence chain: \textit{Entity Existence ($t_1$)} + \textit{Historical Context ($t_2$)}.

\paragraph{3. Active Prevention of Context Bloating.}
The most significant display of ingenuity occurs in \textbf{Turn 3}. The agent retrieves specific confirmation linking "Puebla" to the "Athletic Club." In standard monolithic frameworks (e.g., Search-r1), this new retrieval would typically be appended to the context, causing the input length to grow linearly with search steps.
In contrast, M-ASK employs \textbf{In-Place Refinement}. The Update Agent recognizes that this information is a more precise version of $t_1$. It triggers \texttt{<Update>t1} to refine the query and answer (adding "specifically Puebla") rather than creating a redundant $t_3$. This mechanism maintains a concise state space ($\mathcal{K}_3$), effectively solving the "unconstrained output length" challenge and ensuring the final reasoning is performed on a high-density, noise-free context.

\begin{table*}[p]
\centering
\scriptsize
\renewcommand{\arraystretch}{1.25}
\definecolor{statebg}{gray}{0.95}
\definecolor{toolbg}{rgb}{0.94, 0.97, 1.0}
\definecolor{successgreen}{rgb}{0.0, 0.5, 0.0}

\vspace{0.2cm}

\begin{tabularx}{\textwidth}{lX}
\toprule
\textbf{Target Question:} & What Mexican city had an Athletic Club whose football team played in the Primera Fuerza in 1943? \\
\textbf{Golden Answer:} & \textbf{Puebla} \\
\bottomrule
\end{tabularx}

\vspace{0.1cm}

\begin{tabularx}{\textwidth}{c|l|X}
\toprule
\textbf{Turn} & \textbf{Role} & \textbf{Full Action Output \& Structured Knowledge State ($\mathcal{K}_t$)} \\
\midrule

\textbf{Init} 
& \textbf{Planning Agent} & \texttt{<q1>} What Mexican city had an Athletic Club whose football team played in the Primera Fuerza in 1943? \texttt{</q1>} \\
& & \texttt{<a1>} Guadalajara \texttt{</a1>} \\
& & \texttt{<predicted\_answer>} Guadalajara \texttt{</predicted\_answer>} \\
\hline
& \multicolumn{2}{l}{%
\cellcolor{statebg}%
\begin{minipage}{0.95\textwidth}
\textbf{State $\mathcal{K}_0$ (Initial Structure):} \\
\texttt{"question":} What Mexican city had an Athletic Club whose football team played in the Primera Fuerza in 1943? \\
\texttt{"thinking\_trajectory":} [ \\
\quad $[\tau_1]$ $\langle q_{sub}^{(1)} \rangle$: What Mexican city had an Athletic Club whose football team played in the Primera Fuerza in 1943? \\
\quad \phantom{$[\tau_1]$} $\langle a_{sub}^{(1)} \rangle$: \textcolor{red}{\textbf{Guadalajara}} \textit{(Parametric Hallucination)} \\
] \\
\texttt{"predicted\_answer":} \textcolor{red}{\textbf{Guadalajara}}
\end{minipage}} \\
\midrule

\textbf{Turn 1} 
& \textbf{Search Agent} & \texttt{<search>} What Mexican city had an Athletic Club whose football team played in the Primera Fuerza in 1943? \texttt{</search>} \\
& \cellcolor{toolbg}\textit{Search Engine} & \cellcolor{toolbg}\textit{[External Tool] Retrieved 5 document fragments ($\sim$100 words each) containing keyword matches for "Mexican city", "Athletic Club", "Primera Fuerza", "1943".} \\
& \textbf{Summary Agent} & \texttt{<evidence>} Puebla, which has a city of Puebla, had an Athletic Club, Puebla A.C., whose football team played in the Primera Fuerza. \texttt{</evidence>} \\
& \textbf{Update Agent} & \texttt{<Update>$\tau_1$</Update>} (\textbf{Corrects Hallucination}) \\
& \textbf{Answer Agent} & \texttt{<predicted\_answer>} Puebla \texttt{</predicted\_answer>} \\
\hline
& \multicolumn{2}{l}{%
\cellcolor{statebg}%
\begin{minipage}{0.95\textwidth}
\textbf{State $\mathcal{K}_1$ (Corrected Answer):} \\
\texttt{"question":} What Mexican city had an Athletic Club whose football team played in the Primera Fuerza in 1943? \\
\texttt{"thinking\_trajectory":} [ \\
\quad $[\tau_1]$ $\langle q_{sub}^{(1)} \rangle$: What Mexican city had an Athletic Club whose football team played in the Primera Fuerza in 1943? \\
\quad \phantom{$[\tau_1]$} $\langle a_{sub}^{(1)} \rangle$: \textcolor{blue}{\textbf{Puebla, which has a city of Puebla, had an Athletic Club, Puebla A.C., whose football team played in the Primera Fuerza.}} \\
] \\
\texttt{"predicted\_answer":} \textcolor{blue}{\textbf{Puebla}} \textit{(Updated via retrieval)}
\end{minipage}} \\
\midrule

\textbf{Turn 2} 
& \textbf{Search Agent} & \texttt{<search>} What is a significant event or fact about Puebla A.C.'s football team's participation in the Primera Fuerza in 1943? \texttt{</search>} \\
& \cellcolor{toolbg}\textit{Search Engine} & \cellcolor{toolbg}\textit{[External Tool] Retrieved 5 document fragments ($\sim$100 words each) related to "Puebla A.C.", "Primera Fuerza participation", "1943 events".} \\
& \textbf{Summary Agent} & \texttt{<evidence>} Puebla A.C.'s football team played in the Primera Fuerza prior to the professionalization and development of the Primera División de México in 1943. \texttt{</evidence>} \\
& \textbf{Update Agent} & \texttt{<Add>$\tau_2$</Add>} (Appends Context) \\
& \textbf{Answer Agent} & \texttt{<predicted\_answer>} Puebla \texttt{</predicted\_answer>} \\
\hline
& \multicolumn{2}{l}{%
\cellcolor{statebg}%
\begin{minipage}{0.95\textwidth}
\textbf{State $\mathcal{K}_2$ (Expanded Context):} \\
\texttt{"question":} What Mexican city had an Athletic Club whose football team played in the Primera Fuerza in 1943? \\
\texttt{"thinking\_trajectory":} [ \\
\quad $[\tau_1]$ $\langle q_{sub}^{(1)} \rangle$: What Mexican city had an Athletic Club whose football team played in the Primera Fuerza in 1943? \\
\quad \phantom{$[\tau_1]$} $\langle a_{sub}^{(1)} \rangle$: Puebla, which has a city of Puebla, had an Athletic Club, Puebla A.C., whose football team played in the Primera Fuerza. \\
\quad , \\
\quad $[\tau_2]$ $\langle q_{sub}^{(2)} \rangle$: \textbf{What is a significant event or fact about Puebla A.C.'s football team's participation in the Primera Fuerza in 1943?} \\
\quad \phantom{$[\tau_2]$} $\langle a_{sub}^{(2)} \rangle$: \textbf{Puebla A.C.'s football team played in the Primera Fuerza prior to the professionalization and development of the Primera División de México in 1943.} \\
] \\
\texttt{"predicted\_answer":} \textbf{Puebla}
\end{minipage}} \\
\midrule

\textbf{Turn 3} 
& \textbf{Search Agent} & \texttt{<search>} What Mexican city, specifically Puebla, had an Athletic Club, Puebla A.C., whose football team played in the Primera Fuerza in 1943? \texttt{</search>} \\
& \cellcolor{toolbg}\textit{Search Engine} & \cellcolor{toolbg}\textit{[External Tool] Retrieved 5 document fragments verifying "Puebla", "Puebla A.C.", "Primera Fuerza 1943".} \\
& \textbf{Summary Agent} & \texttt{<evidence>} Puebla, specifically Puebla, had an Athletic Club, Puebla A.C., whose football team played in the Primera Fuerza in 1943. \texttt{</evidence>} \\
& \textbf{Update Agent} & \texttt{<Update>$\tau_1$</Update>} (\textbf{Refines Existing Step}) \\
& \textbf{Answer Agent} & \texttt{<predicted\_answer>} Puebla \texttt{</predicted\_answer>} \\
\hline
& \multicolumn{2}{l}{%
\cellcolor{statebg}%
\begin{minipage}{0.95\textwidth}
\textbf{State $\mathcal{K}_3$ (Refined Trajectory):} \\
\texttt{"question":} What Mexican city had an Athletic Club whose football team played in the Primera Fuerza in 1943? \\
\texttt{"thinking\_trajectory":} [ \\
\quad $[\tau_1]$ $\langle q_{sub}^{(1)} \rangle$: What Mexican city, \textbf{specifically Puebla}, had an Athletic Club, Puebla A.C., whose football team played in the Primera Fuerza in 1943? \\
\quad \phantom{$[\tau_1]$} $\langle a_{sub}^{(1)} \rangle$: \textbf{Puebla, specifically Puebla, had an Athletic Club, Puebla A.C., whose football team played in the Primera Fuerza in 1943.} \\
\quad , \\
\quad $[\tau_2]$ $\langle q_{sub}^{(2)} \rangle$: What is a significant event or fact about Puebla A.C.'s football team's participation in the Primera Fuerza in 1943? \\
\quad \phantom{$[\tau_2]$} $\langle a_{sub}^{(2)} \rangle$: Puebla A.C.'s football team played in the Primera Fuerza prior to the professionalization and development of the Primera División de México in 1943. \\
] \\
\texttt{"predicted\_answer":} \textbf{Puebla}
\end{minipage}} \\
\midrule

\textbf{Turn 4} & \textbf{Search Agent} & \texttt{<end>} (Termination Triggered) \\
\bottomrule
\end{tabularx}

\vspace{0.1cm}

\begin{tabularx}{\textwidth}{lX}
\toprule
\textbf{Final Prediction:} & \textbf{Puebla} \quad \textcolor{successgreen}{\textbf{(\checkmark Matches Golden Answer)}} \\
\bottomrule
\end{tabularx}

\caption{Full execution log for the query: \textit{"What Mexican city had an Athletic Club whose football team played in the Primera Fuerza in 1943?"}. The \textbf{State} rows explicitly show the three components of $\mathcal{K}_t$. The trajectory steps are indexed as $[\tau_i]$, allowing the Update Agent to target specific steps (e.g., \texttt{<Update>$\tau_1$</Update>}).}
\label{tab:case_study_full}

\end{table*}

\end{document}